\documentclass[letterpaper]{article} % DO NOT CHANGE THIS
\usepackage{aaai24}  % DO NOT CHANGE THIS
\usepackage{times}  % DO NOT CHANGE THIS
\usepackage{helvet}  % DO NOT CHANGE THIS
\usepackage{courier}  % DO NOT CHANGE THIS
\usepackage[hyphens]{url}  % DO NOT CHANGE THIS
\usepackage{graphicx} % DO NOT CHANGE THIS
\usepackage{caption} % DO NOT CHANGE THIS AND DO NOT ADD ANY OPTIONS TO IT
\usepackage{algorithm}
\usepackage{algorithmic}

\usepackage{amsmath}
\usepackage{multirow}
\usepackage{amssymb}
\usepackage{bbding}
\usepackage{colortbl}
\definecolor{LightCyan}{rgb}{0.95,0.95,0.95}
\usepackage{newfloat}
\usepackage{listings}
\DeclareCaptionStyle{ruled}{labelfont=normalfont,labelsep=colon,strut=off} % DO NOT CHANGE THIS
\floatstyle{ruled}
\newfloat{listing}{tb}{lst}{}
\floatname{listing}{Listing}
\title{Sunshine to Rainstorm: Cross-Weather Knowledge Distillation \\for Robust 3D Object Detection}
\author{
    Xun Huang \textsuperscript{\rm 1}\equalcontrib,
    Hai Wu \textsuperscript{\rm 1}\equalcontrib,
    Xin Li \textsuperscript{\rm 2},
    Xiaoliang Fan \textsuperscript{\rm 1},
    Chenglu Wen \textsuperscript{\rm 1}\footnote{ Corresponding author.},
    Cheng Wang \textsuperscript{\rm 1}
}
\affiliations{
    \textsuperscript{\rm 1} Xiamen University, 
    \textsuperscript{\rm 2} Texas A \& M University\\
    
    \{huangxun,wuhai\}@stu.xmu.edu.cn, \{clwen, fanxiaoliang, cwang\}@xmu.edu.cn, xinli@tamu.edu
}

\begin{document}

\maketitle

\begin{abstract}
LiDAR-based 3D object detection models inevitably struggle under rainy conditions due to the degraded and noisy scanning signals.
Previous research has attempted to address this by simulating the noise from rain to improve the robustness of detection models. 
However, significant disparities exist between simulated and actual rain-impacted data points. 
In this work, we propose a novel rain simulation method, termed DRET, that unifies \textbf{D}ynamics and \textbf{R}ainy \textbf{E}nvironment \textbf{T}heory to provide a cost-effective means of expanding the available realistic rain data for 3D detection training. 
Furthermore, we present a \textbf{S}unny-to-\textbf{R}ainy \textbf{K}nowledge \textbf{D}istillation (\textbf{SRKD}) approach to enhance 3D detection under rainy conditions.  
Extensive experiments on the Waymo-Open-Dataset show that, when combined with the state-of-the-art DSVT model and other classical 3D detectors, our proposed framework demonstrates significant detection accuracy improvements, without losing efficiency.
Remarkably, our framework also improves detection capabilities under sunny conditions, therefore offering a robust solution for 3D detection regardless of whether the weather is rainy or sunny.

%For a long time, LiDAR-based 3D object detection models have suffered from rainy weather due to degraded and noisy scanning signals. 
%Pioneer research focuses on simulating the rainy noisy points to enhance the detector's robustness. 
%Despite great process has been made, huge differences exist between the simulated points and the real points. 
%We propose a more realistic rain simulation method, termed DRET, for better 3D detection in rainy scenes. DRET incorporates both \textbf{D}ynamics and \textbf{R}ainy \textbf{E}nvironment \textbf{T}heory, cost-effectively expanding the realistic rain data for 3D detection training. 
%Further, we also propose a \textbf{S}unny-to-\textbf{R}ainy \textbf{K}nowledge \textbf{D}istillation (\textbf{SRKD}) tailored for enhanced 3D detection in rain. 
%We combine the proposed framework with the state-of-the-art DSVT model and other classical 3D detectors. 
%Extensive experiments on the Waymo-Open-Dataset large-scale dataset show the ability of SRKD  to substantially improve rainy performance without losing efficiency. 
%Remarkably, our framework also benefits sunny condition detection, providing a new solution for robust 3D detection under both rainy and sunny weather conditions.
\end{abstract}

\section{Introduction}
Recent years have witnessed growing research interest in 3D object detection utilizing point cloud data. For widespread application in autonomous driving, such models must demonstrate robust performance under diverse conditions. Despite progress in 3D object detection on benchmark datasets, achieving consistent and reliable performance, particularly under adverse weather conditions (e.g., rain) remains an open challenge.

\begin{figure}[t]
\centering
\setlength{\belowcaptionskip}{-0.5cm}
\includegraphics[width=\columnwidth]{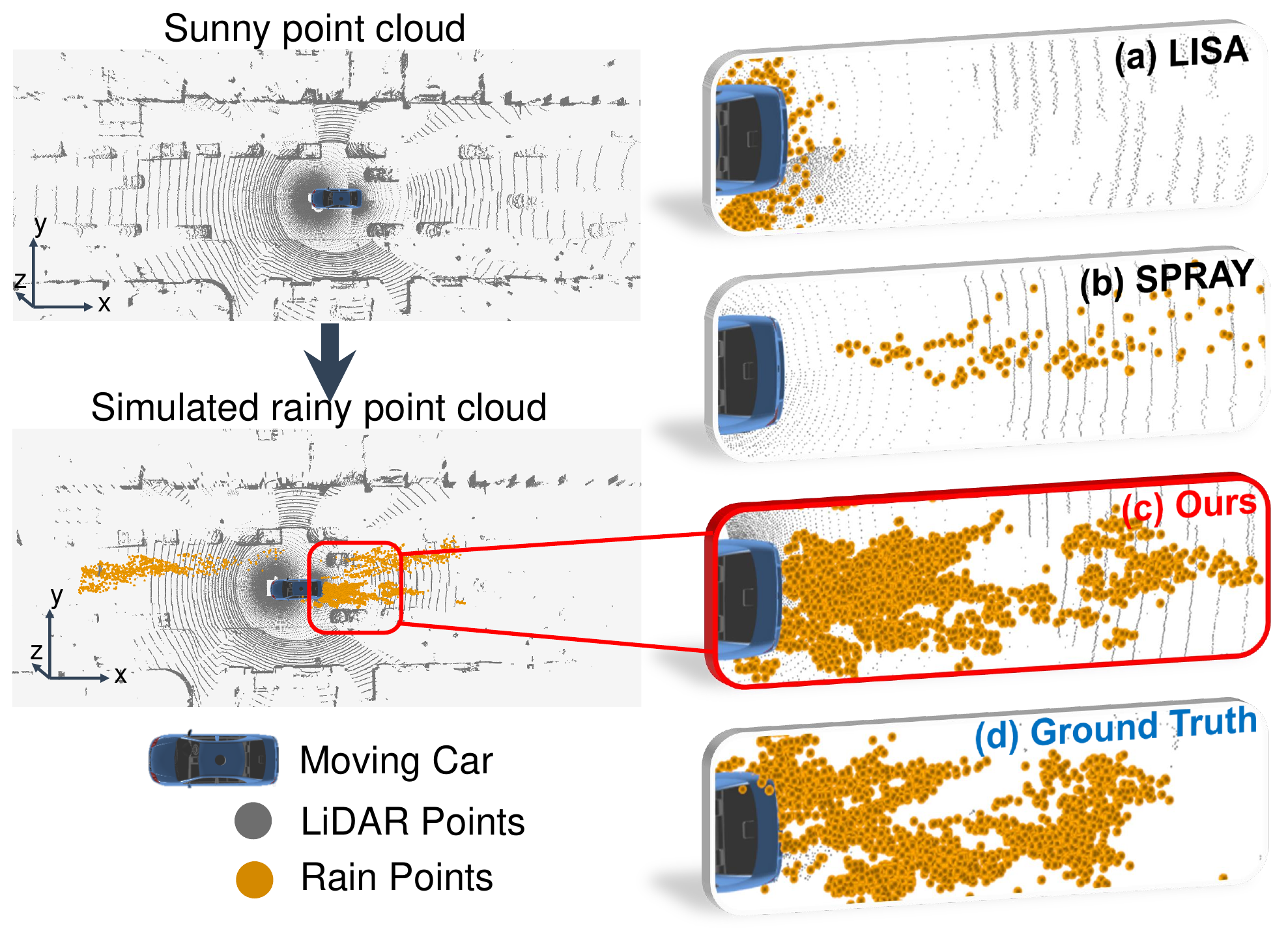} % Reduce the figure size so that it is slightly narrower than the column. Don't use precise values for figure width.This setup will avoid overfull boxes.
\caption{Visualization of rain noise points in groundtruth and different rain simulations.}
\label{fig_intro1}
\end{figure}

Unfortunately, 3D object detection research under rainy weather presents significant challenges at both the data and methods levels. At the data level, a pressing issue is data scarcity. Extant datasets exhibit highly limited rainy samples owing to the high annotation and collection costs. For instance, merely 0.6\% of samples in the Waymo Open Dataset (WOD) \cite{waymo} perception subset in rain. This shortage seriously hinders the research on 3D object detection under rainy weather. The low quality of rainy data obstructs 3D detectors. Analysis of rainy point clouds in the WOD reveals two critical phenomena:
1) \emph{Dense rain noise.}  As LiDAR light pulses cannot penetrate water particles \cite{fogsim}, resulting in noise from water droplets generated by moving vehicles.
2) \emph{Missing points.} Atmospheric parameters like attenuation substantially differ in rain. Thus, many points fall below the LiDAR intensity threshold and are missed.
These phenomena create a considerable domain gap between sunny and rainy data. Robust rainy 3D object detection needs to address both data and model challenges.

Prior work, including LISA \cite{LISA} and SPRAY \cite{spray}, has sought to address the challenge of insufficient training data by simulating adverse weather conditions. 
While these approaches are valuable, they overlook the rainy environmental theory and object-scene dynamics, leading to limitations in the realism of simulated rain. For example, as depicted in Fig.~\ref{fig_intro1}, a significant gap exists between simulated and real rain points, especially in the case of LISA. SPRAY also fails to simulate the phenomenon of missing points that usually occur in rainy scenes. 

%Even given sufficient data, 3D object detection in rain remains challenging due to the gap between sunny and rainy data. Thus, 
In addition to creating realistic simulations, the effective utilization of simulated data is also crucial. Many existing methods~\cite{sim1,fogsim,snowsim} simply augment training data with simulations, but this approach often falls short in adapting detectors effectively to new rainy environments. 
LDNet~\cite{sr} designed an innovative idea of knowledge distillation from sunny to severe weather conditions.
Without a realistic rain simulation, this concept cannot be effectively applied. Moreover, LDNet also overlooks the data disparities between different weather conditions. Only by fully addressing the gaps between rainy and sunny weather can models achieve robustness across diverse weather conditions.

\begin{figure}[!t]
\centering
\includegraphics[width=\columnwidth]{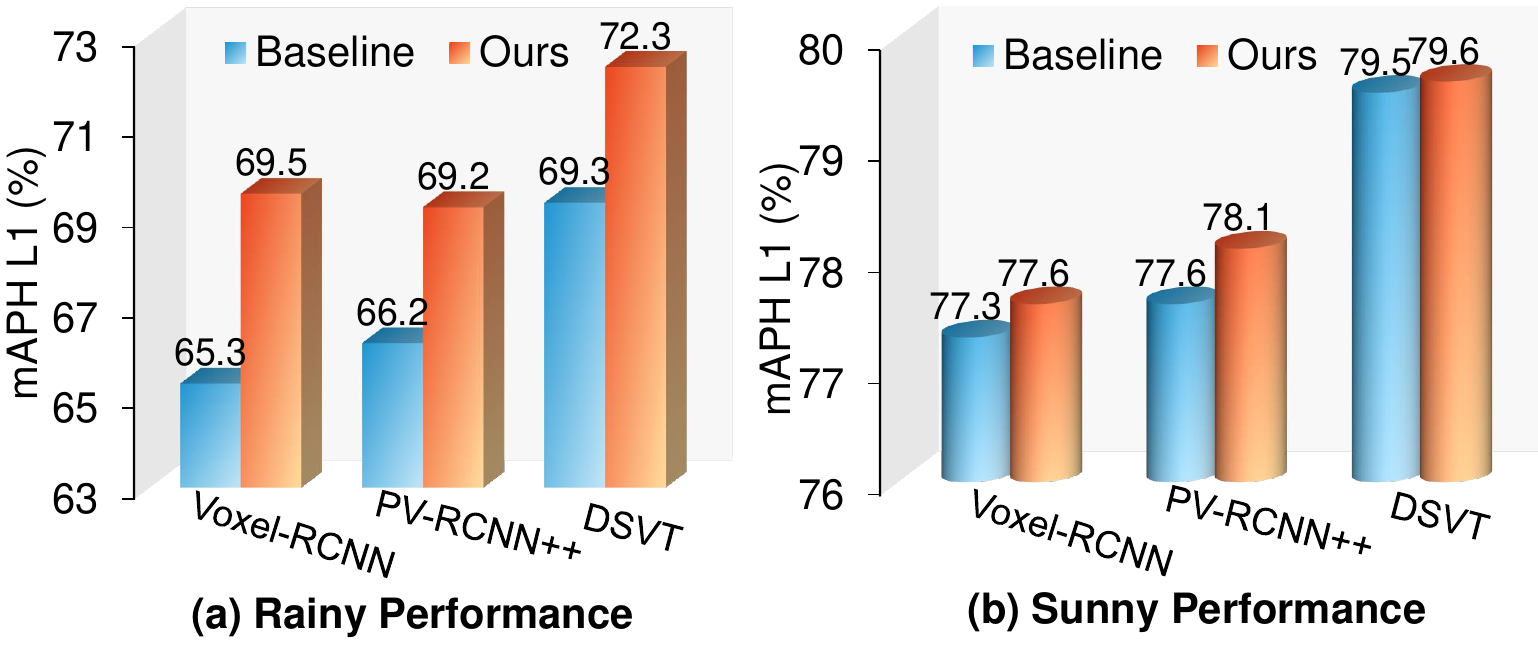} % Reduce the figure size so that it is slightly narrower than the column. Don't use precise values for figure width.This setup will avoid overfull boxes.
\caption{Comparison of the performance under both sunny and rainy weather conditions on the WOD. }
\label{fig_intro2}
\end{figure}

To enhance the robustness of models in rain, we propose innovative approaches at both the data and model levels. We introduce DRET, a method that unifies \textbf{D}ynamics and \textbf{R}ainy \textbf{E}nvironment \textbf{T}heory to create realistic rain data simulations. 
We also present SRKD, a \textbf{S}unny-to-\textbf{R}ainy \textbf{K}nowledge \textbf{D}istillation framework specifically tailored for 3D detectors in rainy conditions. SRKD trains a student network to learn from a sunny teacher detector, distilling knowledge of the same scene under rainy weather. 
Additionally, we design a noise-aware prediction correction module to effectively handle the noise associated with rain. 

Our framework is highly adaptable and can be easily integrated with various 3D detectors. 
As shown in Fig.\ref{fig_intro2}, 
extensive experiments on the WOD validated that DRET and SRKD enhance the rain robustness of all 3D detectors without sacrificing efficiency.
Remarkably, they also improve detectors' performance under sunny conditions, possibly due to their increased robustness to sparse objects. 

Our contributions are : 
(1) We analyzed the impact of various phenomena on 3D object detection in rainy weather, and proposed a new method, DRET, for more realistic rain simulation in 3D object detection. 
(2) 
We designed SRKD, a generalized framework to address the challenges posed by weather disparities in straightforward sunny-to-rainy distillation. It can significantly improve the robustness of the 3D detector in rain without compromising efficiency.
(3) We validated the effectiveness and universality of our proposed approach through extensive testing on the state-of-the-art model, DSVT~\cite{dsvt}, as well as other classical 3D detectors, including PV-RCNN++ and Voxel-RCNN.

%(3) Demonstrating the efficacy and generality of our proposed approach through extensive experiments on state-of-the-art model DSVT and other classical 3D detectors including PV-RCNN++ and Voxel-RCNN.

\section{Related Work}
\paragraph{Simulation of rain.} Physics-based simulations~\cite{sim1,snowsim,fogsim, LISA} have been explored to reproduce point clouds under adverse weather conditions such as snow and fog. These methods help alleviate the issue of data scarcity to some extent, but they are not fully effective in simulating rainy weather. For instance, LISA~\cite{LISA} employs a rain simulation algorithm that relies on LiDAR light scattering augmentation, but it overlooks the impact of dense rain noise in rainy conditions. SPRAY~\cite{spray} uses dynamics to simulate the water splashing effect caused by vehicles in the rain, but it fails to accurately mimic the actual distribution of splashed water due to the absence of a well-grounded physical theory. Consequently, substantial discrepancies persist between these simulated rainy datasets and actual rainy conditions.

\paragraph{3D object detection.} 
Current 3D object detection methods primarily focus on clear weather conditions and can be categorized into single-stage and two-stage approaches. Single-stage methods, such as SECOND  \cite{second}, PointPillars  \cite{voxelrcnn}, SA-SSD  \cite{sassd}, and SE-SSD  \cite{sessd}, utilize voxel-based sparse convolution or point-based set abstraction for feature extraction. In two-stage algorithms, Voxel-RCNN  \cite{voxelrcnn}, SFD  \cite{sfd}, and VirConv  \cite{virconv} employ voxel-based sparse convolution, while PointRCNN  \cite{pointrcnn} and STD  \cite{std3d} use point-based set abstraction. Notably, PV-RCNN  \cite{pvrcnn}, CT3D  \cite{ct3d}, and PV-RCNN++  \cite{pvrcnn++} combine voxel-based and point-based operations in the two-stage approach. DSVT  \cite{dsvt} introduces an efficient and deployable 3D transformer backbone that achieves state-of-the-art performance on WOD. Knowledge distillation has been explored in 3D object detection, including SparseKD  \cite{spareKD}, which examines lightweight models, and other works  \cite{boost1,boost2} attempting distillation from multimodal/multi-frame models to single-modal/single-frame models. However, these mainstream methods still lack robustness in rain.

\paragraph{3D object detection in adverse weather.} 
Recent studies, such as \cite{denoise_snow,cnn_denoise}, have introduced lightweight approaches that employ techniques like semantic segmentation or filtering algorithms to remove dense noise, thereby minimizing its impact on detection models.
However, these methods heavily rely on the denoising model's performance, leading to unstable detection results, and they do not account for the issue of missing points in rainy weather. \cite{spg} designed a general completion framework that addresses the problem of domain adaptation across different weather and environmental conditions. It achieves promising improvements on the WOD domain adaptation dataset, which contains a significant amount of rainy data. However, it solely focuses on the phenomenon of missing points, disregarding the impact of other phenomena.

\begin{figure*}[!t]
\centering
\includegraphics[width=\textwidth]{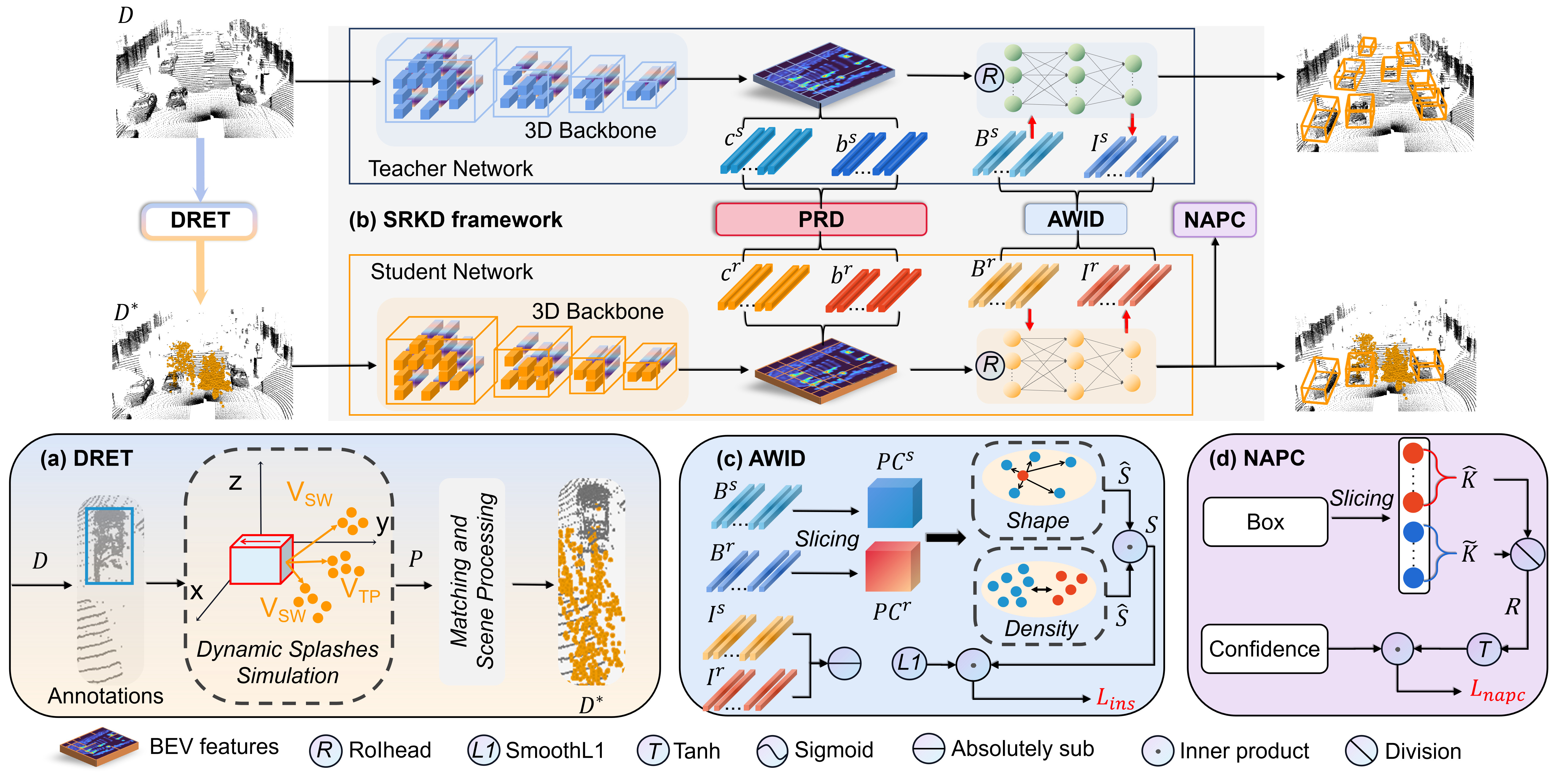}
%\vspace{-12mm}
\caption{The overview of our method, including (a) DRET for rain simulation and (b) SRKD framework for 3D object detection. DRET involves a two-stage process. The first stage simulates the dynamic splashes to generate rain particles, and the second stage matches rain particles with the point cloud and processes the scene based on the rainy environment theory. SRKD enables sunny-to-rainy knowledge distillation with the help of (c) AWID and PRD, which avoids the problems associated with the weather domain gap. PRD is a classic response distillation with some adjustments, which will be introduced in the SRKD method. Additionally, the (d) NAPC module is designed to mitigate the influence of rain-induced noise.}
\label{frame2}

\end{figure*}

\section{Method}\label{sim}

The challenges associated with data limitations and model adaptivity create significant obstacles for robust object detection in rainy conditions. Therefore, this work proposes two main approaches: the simulation of realistic rain and new object detection tailored for rainy environments. 
As illustrated in Fig.~\ref{frame2}, we introduce DRET (a), a method that generates more accurate and realistic simulated rain data for detector training. Additionally, we present SRKD (b), a general framework that enhances the detector's robustness in rainy conditions without compromising efficiency.

\subsection{DRET Rain Simulation Method}
Rain simulation models take a sunny point cloud data $D$ as input and generate a simulated rainy point cloud $D^*$. 
Existing simulation models such as SPRAY \cite{spray} and LISA \cite{LISA} have not considered both the dynamic particle simulation and rainy environment theory in a unified pipeline. 
We propose a new integrated, two-stage simulation method.  
As depicted in Fig.~\ref{frame2} (a), its first stage employs the Unity3D engine to simulate dynamic splashing. It also utilizes Perlin noise \cite{perlin} to replicate wind interference. However, this particle simulation alone doesn't account for LIDAR reflection intensity for rain particles. Thus, we additionally introduce a second stage of performing a scene process following the rainy environment theory \cite{LISA, fogsim} to address this limitation.

%To get real rain particles, SPRAY works with the Unity3D engine to simulate particle motion. That is an excellent thought, thus 
In the first stage, similar to SPRAY~\cite{spray}, we utilize the particle emitter in the dynamics particle system to simulate water splashes caused by moving vehicles. We incorporate three splash mechanisms: bow wave (BW), side wave (SW), and tread pickup (TP) following SPRAY. 
But unlike SPRAY, we use Perlin noise to introduce random accelerations along the $x$, $y$, and $z$ axes. 
This extra design better imitates wind perturbations and contributes to a broader range of realistic rain particle behaviors. 

After the initial stage, we obtain sets of rain particles. However, merging them directly into the sunny point clouds, as done in SPRAY, is not a good idea. Because it results in the absence of LiDAR intensity information for rain particles, and the intensity of the original points would remain unadjusted for rainy conditions. Therefore, we introduce an additional scene processing step based on the rainy environment theory in the second stage.

Before conducting the scene processing, we collect the location, number, inclination and maximum distance parameters of the LiDARs used in the dataset. These parameters are used to establish a correspondence between the point cloud data $D$ and the rain particle set $P$. This gives us matched pairs $\{(D_i, P_j) \} $, in which each pair of points originates from the same laser beam reflection.  %$\mid \text{match}(D_i, P_j) \text{ is true}\}$. Here, $\text{match}$ indicates whether the two points reflected from the same laser beam. 

Next, unlike LISA's random generation of rain particles, we employ more realistic rain particle sets $P$ for scene processing. We also adopt more accurate formulas from fog simulation \cite{fogsim}. 
Specifically, for a matched pair $(D_i, P_j)$, where $D_i$ is a point with reflection intensity $I_i$ and distance $R_i$, and $P_j$ is a rain particle with distance $R_j$, we calculate the intensity $I_j$ of $P_j$ using 

\begin{equation}
\begin{aligned}
I_j= \enspace & C_A P_0 \beta \int_0^{2 \tau_H} \sin ^2\left(\frac{\pi}{2 \tau_H} t\right) \frac{\exp \left(-2 \alpha\left(R_j-\frac{c t}{2}\right)\right)}{\left(R_j-\frac{c t}{2}\right)^2} \times \\
& \times \gamma\left(R_j-\frac{c t}{2}\right) U\left(R_i-R_j+\frac{c t}{2}\right) d t,
\end{aligned}
\end{equation}
where $R_1, R_2$ are the sensor optical parameters, $\beta_0$ is the differential reflectance of point, $\beta$ is the scattering rate, $\alpha$ is the attenuation coefficient, $c$ is the speed of light, $\tau_H$ is the half-power pulse width, $U$ is the Heaviside function, and 
\begin{equation}
C_AP_0 = I_i\frac{R_i^2}{\beta_0},
\gamma(R)=\left\{\begin{array}{cl}
0, & R \leq R_1 \\
\frac{R-R_1}{R_2-R_1}, & R_1 \textless R \textless R_2 \\
1, & R_2 \leq R
\end{array}\right..
\nonumber
\end{equation} 
The calculation is performed using Simpson’s $^1/_3$ rule for numerical integration.

After calculating the intensity $I_j$ of $P_j$, we use $P_j$ to replace $D_i$ in $D$ to simulate the occlusion of dense noise in real rain. This process yields a new point cloud $D'$.
Then, following LISA \cite{LISA}, we adjust the part of original sunny points in $D'$ to reflect rainy weather conditions. We collect points where the LiDAR received power $\mathcal{P}$ exceeds the minimum LiDAR received power $\mathcal{P}_{min}$ to get the simulated rainy point cloud $D^*$: 
\begin{equation}
D^* = \{{D_i'} \mid \mathcal{P}_i \geq \mathcal{P}_{min}\}.
\end{equation}
where $\mathcal{P}_i = \frac{I_{i}}{R_i^2}$, $\mathcal{P}_{min} = 0.9R_{max}^2$, and $R_{max}$ is the max distance of LiDAR.

\iffalse
Then we modify the part of original sunny points in $\tilde{PC}$ to rainy weather. For a original sunny point $\tilde{PC}_i$ in $\tilde{PC}$ with intensity $I_i$ at distance $R_i$, the fixed intensity under rainy weather $\tilde{I_{i}}$ is obtained according to the attenuation coefficient $a$:
\begin{equation}
\tilde{I_{i}} = I_i\mathit{\exp}\left({-2 \alpha R_{i}}\right),
\end{equation}
and its LiDAR received power ${\mathcal{P}_i}$ in rain is:
\begin{equation}
\mathcal{P}_i = \frac{\tilde{I_{i}}}{R_i^2},
\end{equation}
meanwhile, the received minimum power threshold $\mathcal{P}_{min}$ can be calculated by the max distance $R_{max}$ of LiDAR:
\begin{equation}
\mathcal{P}_{min} = 0.9R_{max}^2,
\end{equation}
finally, we collect those points in $\tilde{PC}$ where $\mathcal{P}$ is higher than $\mathcal{P}_{min}$ to get the simulated rainy point cloud $PC^*$: 
\begin{equation}
PC^* = \{\tilde{PC}_i \mid \mathcal{P}_i \geq \mathcal{P}_{min}\}.
\end{equation}
\fi

\begin{figure}[!t]
\centering
\includegraphics[width=1.0\columnwidth]{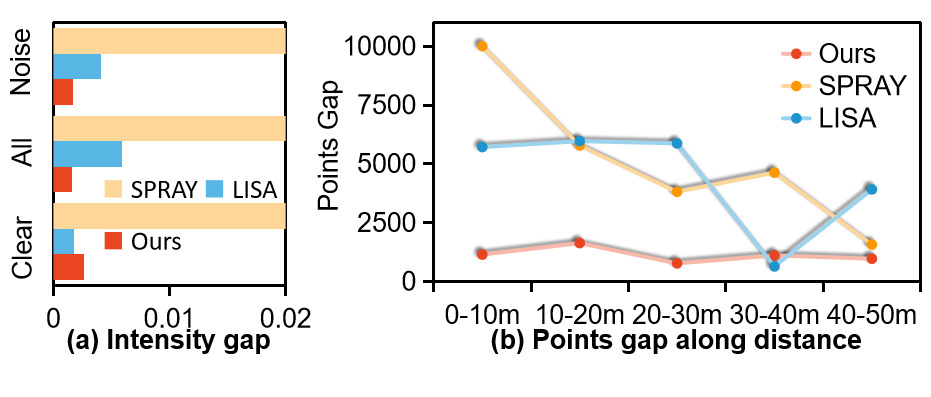} % Reduce the figure size to slightly narrower than the column. Don't use precise values for figure width. This setup will avoid overfull boxes.
\caption{The average intensity gap. (b) The average points gap in each distance interval.}
\label{datadist}
\end{figure}

\paragraph{Comparison with other rain simulation methods.} 
We simulated 10k rainy point cloud data using 
LISA~\cite{LISA}, 
SPRAY~\cite{spray}, and our DRET method, respectively. 
We then analyzed the average intensity gap between these simulated data and real rain data (see Fig.~\ref{datadist} (a)). 
The SPRAY has a serious gap in intensity with the real data, mainly because of the missing rain particle intensity calculations and original clear point corrections, as mentioned earlier. Although LISA seems to have a smaller intensity gap in the clear points this is a fluctuation caused by the randomization of the rain rate parameter. The intensity gap of our method is significantly less than LISA in terms of noise and all points, especially on all points.
By analyzing the average points gap along distance (see Fig.~\ref{datadist} (b)), we can also find that SPRAY has a huge gap because it can not filter the low received power points. While LISA has a small points gap at 30-40m, the overall points gap fluctuates greatly and the overall gap is significantly larger than our method.
In contrast, our simulated data is smaller in both the intensity gap and points gap, indicating that our method is more realistic.

In summary, our DRET simulation method can offer a lot of realistic simulated rain data, thereby broadening the possibilities for research into robust 3D object detection in rain.

\subsection{Sunny-to-Rainy Knowledge Distillation}

Expanding rainy data using DRET as a data augmentation during training is valuable.
However, this strategy alone does not address the gaps existing between sunny and rainy data. 
Even applying distillation techniques as seen in previous works~\cite{boost1, boost2, sr} cannot effectively bridge the weather-related gaps. 
In light of this, we undertake an initial analysis of how these weather disparities impact 3D object detection. 
Based on these analyses, we design the SRKD framework to overcome the challenges posed by the weather gaps when distilling knowledge from sunny to rainy conditions.
As shown in Fig.\ref{frame2} (b), the SRKD framework brings the following technical contributions. 
Firstly, we devise a knowledge distillation approach capable of facilitating the transition of 3D detectors from sunny to rainy scenarios. This includes the integration of \emph{Adaptively Weighted Instance Distillation (AWID)} and \emph{Precise Response Distillation (PRD)}. Additionally, we introduce \emph{Noise-Aware Prediction Correction (NAPC)} to mitigate the influence of rain-induced noise.

\begin{figure}[!t]
\centering
\includegraphics[width=1.0\columnwidth]{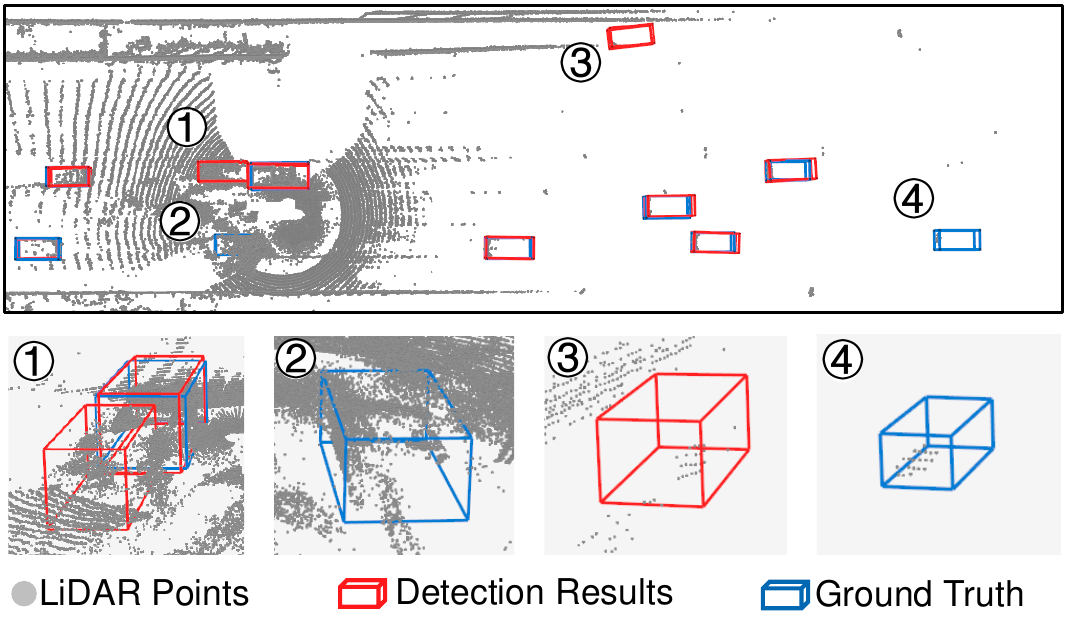} % Reduce the figure size to slightly narrower than the column. Don't use precise values for figure width. This setup will avoid overfull boxes.
\caption{Examples of rainy condition effects on 3D object detection. The blue and red boxes represent groundtruths and predictions.}
\label{detect_rain}
\end{figure}

\paragraph{Analysis on impact of rain on 3D object detection.} 
As previously discussed, the presence of dense rain noise and the absence of points in rainy conditions pose significant challenges for 3D object detection. Table \ref{table:motivation} highlights the substantial decrease in both precision and recall for Voxel-RCNN \cite{voxelrcnn} on the WOD dataset during rainy conditions. This effect is further illustrated in Fig.~\ref{detect_rain}, which depicts the impact of rainy weather on the detector's performance: 
1) \emph{Dense rain noise.} The presence of dense rain noise often results in numerous false positives (Fig.~\ref{detect_rain}\textcircled{1}). 
Additionally, these noise induced detections can erod the confidence of predictions, potentially leading to false negatives (Fig.\ref{detect_rain}\textcircled{2}).
2) \emph{Missing points.} The sparse clustered background points due to rain induce false positives  (Fig.\ref{detect_rain}\textcircled{3}). Meanwhile, the absence of foreground points creates sparse point representations for many objects, making them difficult to recall accurately(Fig.\ref{detect_rain}\textcircled{4}).

\begin{table}[!t]

    \small
	\centering
	\begin{tabular}{c | c  c   c  | c   c  c }
		\hline
		\multirow{2}*{Weather}  &\multicolumn{3}{c|}{Precision}   &\multicolumn{3}{c}{Recall} \\  
		 &\normalsize0.3 &\normalsize0.5 &\normalsize0.7  &\normalsize0.3 &\normalsize0.5 &\normalsize0.7\\
		\hline
		Sunny     &\normalsize{46.8}	&\normalsize{44.8}	&\normalsize{34.6}	&\normalsize{87.3}	&\normalsize{84.1}	&\normalsize{69.7}\\
        Rainy     &\normalsize33.6	&\normalsize32.3	&\normalsize27.3	&\normalsize83.2	&\normalsize80.0	&\normalsize66.7\\
        \textit{Degradation}     &\textit{-13.2}	&\textit{-12.5} &\textit{-7.3} &\textit{-4.1} &\textit{-4.1} &\textit{-3.0}\\
		\hline
	\end{tabular}
    \caption{Precision and recall of vehicle in different weather. 0.3, 0.5, and 0.7 are the IoU thresholds. }

    \label{table:motivation}
\end{table}

Thus, instead of naively utilizing simulated data as augmentation, it is crucial to propose a novel framework to address the domain gap between rainy and sunny conditions.

\paragraph{AWID.} 
Distilling sunny instances into rainy instances can help students efficiently extract features from indistinguishable and sparse objects. However, directly distilling instance features from sunny teachers to rainy students, following \cite{boost1, boost2}, poses a great challenge due to the gap between sunny and rainy object variations.
By comparing and analyzing the data of instances under rainy and sunny conditions, we found that the changes in instances are mainly reflected in differences in density and shape. To overcome this challenge, our proposed AWID utilizes the similarity between corresponding sunny and rainy objects as distillation weights for instance features. Our adaptive weighting reduces the domain gap caused by variations in density and shape under different weather conditions. The similarity weight consists of two components: density similarity and shape similarity. For a groundtruth object box $B_i$ in sunny and rainy conditions, with point clouds $PC_i^s$ and $PC_i^r$ of $d_i^s$ and $d_i^r$ points, the density similarity ${\hat{\mathcal{S}}_i}$ is:
\begin{equation} \label{eq3}
 { \hat{\mathcal{S}_i} }= tanh \left(\frac{min \left(d_i^s, d_i^r\right)}{|d_i^s-d_i^r|+\epsilon }\right),
\end{equation}
$\epsilon$ = 1e-6 prevents division by zero. Then calculate the shape similarity $ { \tilde{\mathcal{S}_i} }$ by Chamfer Distance $d_{CD} $ \cite{charmfer}:
\begin{equation} \label{eq4}
 {{ \tilde{\mathcal{S}_i}}=1 - tanh \left(d_{CD}\left(PC_i^s,PC_i^r\right)\right)}.
\end{equation}
Combining Eqs.\ref{eq3} and Eqs.\ref{eq4} we obtain the similarity $\mathcal{S}_i$:
\begin{equation}
 {{ \mathcal{S}_i }=\hat{\mathcal{S}_i} \cdot \tilde{\mathcal{S}_i}}.
\end{equation}
So we can transfer instance feature knowledge for each box from sunny to rainy by using similarity  ${\mathcal{S}}$ as weight:
\begin{equation}
\mathcal{L}_{ins}=\frac{1}{|B|} \sum_{i \in B}  \mathcal{S}_i \cdot \mathcal{L}_{sml1}\left(I_{i}^s,I_{i}^r\right),
\end{equation}
where $B$ is the groundtruth boxes; $I_i$ is the $i$-th instance feature, superscript $s, r$ represent sunny and rainy.

\paragraph{PRD.} 
Our AWID greatly aids in sunny-to-rainy knowledge distillation. However, it remains specific to the RoI head in two-stage works and lacks comprehensiveness. And we expect the model to predict  consistent results for the same scene on sunny and rainy weather.

To address the remaining challenges, we introduce response distillation as a classical distillation approach \cite{spareKD}. This approach promotes prediction consistency between sunny teachers and rainy students. We focus on the teacher's output in positions where the confidence exceeds the threshold $\mathcal{T}$ (0.5), aiming to prevent negative effects from already difficult sunny objects. Specifically, at estimated foreground positions $G$:
\begin{equation} \label{eq1}
\begin{split}
 \mathcal{L}_{cls}^{r} = \frac{1}{|G|} \sum_{i \in G} \mathbb{I}\left(\phi\left(c_{i}^s\right) \geq \mathcal{T}\right) \cdot \mathcal{L}_{mse} \left(c_{i}^{r}, c_{i}^s\right),
\end{split}
\end{equation}
\begin{equation} \label{eq2}
\mathrm{and} \enspace {\mathcal{L}_{reg}^{r}=\mathcal{L}_{reg}^{3D}\left(b^r, b^s\right)},
\end{equation}
where the superscripts $r$ and $s$ indicate rainy student and sunny teacher, respectively; $c$ and $b$ represent classification and box regression predictions; $\phi$ is the sigmoid function; $\mathbb{I}$ filters the teacher's classification output, focusing on high-confidence predictions; $\mathcal{L}_{cls}^r$ and $\mathcal{L}_{reg}^r$ are the classification and regression response distillation losses. By combining Eqs.\ref{eq1} and Eqs.\ref{eq2}, the total precise response distillation loss is obtained as:
\begin{equation}
{\mathcal{L}_{rsp} = \lambda_1{\mathcal{L}_{cls}^{r}}+\lambda_2{\mathcal{L}_{reg}^{r}}}.
\end{equation}
where set $\lambda_1$ = 15, $\lambda_2$ = 0.2 according to  \cite{spareKD}.

\begin{table*}[!ht]
	\centering
    \resizebox{\textwidth}{!}{
    
	\begin{tabular}{c|c|c| c| c | c | c }
        
		\hline
		\multirow{2}*{Methods}   &{All(L1)} &{All(L2)}    &{Vehicle(L1)} &{Vehicle(L2)} &{Ped.(L1)} &{Ped.(L2)}\\             
                                       &mAP/mAPH   &mAP/mAPH
                                       &mAP/mAPH     &mAP/mAPH     
                                       &mAP/mAPH     &mAP/mAPH\\
        \hline               
        Voxel-RCNN \cite{voxelrcnn}
        &65.3/62.6  &55.8/53.7  &73.2/72.8  &66.6/66.2  &57.4/52.4  &45.1/41.1   \\
        + Our DRET-Aug\&SRKD
        &\textbf{69.5}/\textbf{66.2}	&\textbf{60.1}/\textbf{57.3}	&\textbf{75.0}/\textbf{74.6} &\textbf{68.5}/\textbf{68.1}	&\textbf{63.9}/\textbf{57.8}	&\textbf{51.6}/\textbf{46.4} \\
        
        \rowcolor{LightCyan}
        \textit{Improvement}
        
        &\textit{+4.2}/\textit{+3.6}	&\textit{+4.3}/\textit{+3.6}	&\textit{+1.8}/\textit{+1.8}	&\textit{+1.9}/\textit{+1.9}	&\textit{+6.5}/\textit{+5.4}	&\textit{+6.5}/\textit{+5.3} \\
    \hline\hline
        
        PV-RCNN++ \cite{pvrcnn++}
        &66.2/63.4	&56.8/54.5	&73.7/73.2	&67.1/66.7	&58.6/53.5	&46.5/42.2 \\

        + Our DRET-Aug\&SRKD
        &\textbf{69.2}/\textbf{66.5}	&\textbf{59.8}/\textbf{57.6}	&\textbf{76.7}/\textbf{76.2} &\textbf{69.4}/\textbf{69.0}	&\textbf{61.6}/\textbf{56.7}	&\textbf{50.1}/\textbf{46.2} \\

        \rowcolor{LightCyan}
        \textit{Improvement}

        &\textit{+3.0}/\textit{+3.1}	&\textit{+3.0}/\textit{+3.2}	&\textit{+3.0}/\textit{+3.0}	&\textit{+2.3}/\textit{+2.3}	&\textit{+3.0}/\textit{+3.2}	&\textit{+3.6}/\textit{+4.0} \\
    \hline\hline
        
        DSVT \cite{dsvt}
        &{69.3}/{66.5}	&{60.4}/{57.9}	&{75.1}/{74.7}	&{68.4}/{68.0}	&{63.4}/{58.3}	&{52.3}/{47.8} \\
        
        + Our DRET-Aug\&SRKD
         &\textbf{72.7}/\textbf{69.7} &\textbf{63.6}/\textbf{61.1}	&\textbf{76.8}/\textbf{76.4} &\textbf{70.2}/\textbf{69.8}	&\textbf{68.5}/\textbf{63.0} &\textbf{57.0}/\textbf{52.3}\\
        
        \rowcolor{LightCyan}
        \textit{Improvement}
 
        &\textit{+3.4}/\textit{+3.2}	&\textit{+3.2}/\textit{+3.2}	&\textit{+1.7}/\textit{+1.7}	&\textit{+1.8}/\textit{+1.8}	&\textit{+5.1}/\textit{+4.7}	&\textit{+4.7}/\textit{+4.5} \\		
  \hline
	\end{tabular}
    }
    \caption{The rainy testing results of 3D detectors within our proposed DRET-Aug and SRKD on the WOD-DA (100\% training data). All the results are based on our implementation following the open-source code. The best results are in bold. }	
    \label{table:waymorain}
\end{table*}

\begin{table*}[!ht]
	\centering
    \setlength{\tabcolsep}{3pt}
    \resizebox{\textwidth}{!}{
	\begin{tabular}{c|c |c | c | c | c | c  | c | c  }
		\hline
		\multirow{2}*{Methods}    &{All(L1)} &{All(L2)}    &{Vehicle(L1)} &{Vehicle(L2)} &{Ped.(L1)} &{Ped.(L2)} &{Cyc.(L1)} &{Cyc.(L2)}\\          
                                       &mAP/mAPH   &mAP/mAPH
                                       &mAP/mAPH     &mAP/mAPH     
                                       &mAP/mAPH     &mAP/mAPH        
                                       &mAP/mAPH     &mAP/mAPH\\
        \hline               
        Voxel-RCNN
        &{77.3}/{74.9}	&{70.6}/{68.4}	&{78.4}/{77.9}	&{69.8}/{69.4}	&{80.5}/{74.7}	&{71.7}/{66.4}	&{73.1}/{72.0}	&{{70.4}}/{69.3} \\
        
        + Ours  

        &\textbf{77.6}/\textbf{75.2}	&\textbf{71.0}/\textbf{68.6}		&\textbf{78.5}/\textbf{78.0}	&\textbf{70.0}/\textbf{69.5}	&\textbf{81.0}/\textbf{75.1}	&\textbf{72.2}/\textbf{66.7}	&\textbf{73.4}/\textbf{72.4}	&\textbf{70.7}/\textbf{69.7} \\

        \iffalse
        \rowcolor{LightCyan}
        \textit{Improvement}

        &\textit{+0.3}/\textit{+0.3}	&\textit{+0.4}/\textit{+0.2}	&\textit{+0.1}/\textit{+0.1}	&\textit{+0.2}/\textit{+0.1}	&\textit{+0.5}/\textit{+0.4}	&\textit{+0.5}/\textit{+0.3}	&\textit{+0.3}/\textit{+0.2}	&\textit{+0.3}/\textit{+0.4} \\
        \fi
    \hline
        
        PV-RCNN++
        &{77.6}/{75.0}	&{70.8}/{68.6}	&{79.1}/{78.6}	&{70.3}/{69.9}	&{81.1}/{75.0}	&{72.5}/{66.8}	&{72.5}/{71.4}	&{69.7}/{69.2} \\

        + Ours
        &\textbf{78.1}/\textbf{75.6}	&\textbf{71.4}/\textbf{69.0}	&\textbf{79.3}/\textbf{78.8}	&\textbf{70.6}/\textbf{70.1}	&\textbf{81.2}/\textbf{75.2}	&\textbf{72.6}/\textbf{66.9}	&\textbf{73.7}/\textbf{72.7}	&\textbf{70.9}/\textbf{69.9} 
        \\
        \iffalse
        \rowcolor{LightCyan}
        
        \textit{Improvement}

        &\textit{+0.5}/\textit{+0.6}	&\textit{+0.6}/\textit{+0.4}	&\textit{+0.2}/\textit{+0.2}	&\textit{+0.3}/\textit{+0.2}	&\textit{+0.1}/\textit{+0.2}	&\textit{+0.1}/\textit{+0.1}	&\textit{+1.2}/\textit{+1.3}	&\textit{+1.2}/\textit{+0.7} \\
        \fi
    \hline
        
        DSVT
        &{79.5}/{76.9}	&{73.4}/{70.9}	&{79.1}/{78.7}	&{71.3}/{70.8}	&{82.5}/{76.2}	&{75.2}/{69.2}	&\textbf{76.9}/\textbf{75.8}	&{73.8}/{72.8} \\
        + Ours
 
        &\textbf{79.6}/\textbf{77.1}	&\textbf{73.7}/\textbf{71.2}	&\textbf{79.5}/\textbf{79.0}	&\textbf{71.6}/\textbf{71.2}	&\textbf{82.6}/\textbf{76.5}	&\textbf{75.3}/\textbf{69.5}	&{76.8}/{75.7}	&\textbf{74.1}/\textbf{73.0}\\
        \iffalse
        \rowcolor{LightCyan}

        \textit{Improvement}

        &\textit{+0.1}/\textit{+0.2}	&\textit{+0.3}/\textit{+0.3}	&\textit{+0.4}/\textit{+0.3}	&\textit{+0.3}/\textit{+0.4}	&\textit{+0.1}/\textit{+0.3}	&\textit{+0.1}/\textit{+0.3}	&\textit{-0.1}/\textit{-0.1}	&\textit{+0.3}/\textit{+0.2}\\
        \fi
    \hline
	\end{tabular}
    }
    \caption{The sunny testing results of 3D detectors within our proposed DRET-Aug and SRKD on the WOD-P (100\% training data). All the results are based on our implementation following the open-source code. The best results are in bold. }	
    \label{table:waymosun}
\end{table*}

\paragraph{NAPC.} 
The previous modules improve model robustness in rainy conditions via sunny-to-rainy distillation. However, this implicit processing alone cannot mitigate false positives caused by dense rain noise effectively. The rain noise caused by splashing caused by the high speed movement of vehicle tires is heavily distributed around the instance. We leverage the key advantage of simulated rainy data, which provides self-contained rain noise labels for correcting prediction inaccuracies. We introduce a Noise-Aware Prediction Correction head, NAPC, which focuses on suppressing rain noise.
For a predicted box $\mathcal{B}_i$ with $\mathcal{K}$ points, including $\hat{\mathcal{K}}$ noise points and $\tilde{\mathcal{K}}$ non-noise points ($\hat{\mathcal{K}}+\tilde{\mathcal{K}}=\mathcal{K}$), the noise ratio $\mathcal{R}_i$ is calculated as:
\begin{equation}
\mathcal{R}_i = \frac{\hat{\mathcal{K}}}{\tilde{\mathcal{K}}+\epsilon},
\end{equation}
where $\epsilon$ = 1e-6 is a small constant to prevent division by zero. The noise-aware prediction correction loss $\mathcal{L}_{napc}$ is calculated as:
\begin{equation}
\mathcal{L}_{napc}=\frac{1}{|\mathcal{B}|} \sum_{i \in \mathcal{B}} tanh(\mathcal{R}_i) \cdot \mathcal{C}_i,
\end{equation}
where $\mathcal{B}$ represents prediction boxes, and $\mathcal{C}_i$ is the confidence of the $i$-th prediction box.

\paragraph{Overall Loss Function.} 
We train the rainy student while keeping the pre-trained sunny teacher fixed (see Fig.\ref{frame2}) with the following supervision losses and distillation losses:
\begin{equation}
\label{eq_all}
\mathcal{L}=\mathcal{L}_{cls}+\mathcal{L}_{reg}+\eta_1\mathcal{L}_{ins}+\eta_2\mathcal{L}_{rsp}+\eta_3\mathcal{L}_{napc}
\end{equation}
where $\mathcal{L}_{cls}$ is classification loss and $\mathcal{L}_{reg}$ is regression loss; $\eta_1$, $\eta_2$ and $\eta_3$ are hyper-parameters.

\section{Experiments}
We conducted evaluations of our proposed DRET and SRKD on the Waymo Open Dataset (WOD) \cite{waymo}. To demonstrate the effectiveness of 3D detection under rainy condition, we compared the performance of state-of-the-art model DSVT \cite{dsvt} and PV-RCNN++ \cite{pvrcnn++}, Voxel-RCNN \cite{voxelrcnn}.

\paragraph{Dataset.}
The WOD comprises two subsets: Perception (WOD-P) and Domain Adaptation (WOD-DA). The WOD-P contains $\sim$158k training  frames and $\sim$40k validation frames, predominantly sunny conditions (99.4\%). We directly use the WOD-P validation for sunny testing. the WOD-DA includes various weather conditions such as foggy, cloudy, and rainy. For our rainy testing, we selected $\sim$3k rainy  frames for rainy testing.
The training process of all models used only the training set of WOD-P.

\paragraph{Metrics.}
The official metrics of WOD are mean Average Precision (mAP(L1), mAP(L2)) with Heading (mAPH(L1), mAPH(L2)), where L1 and L2 denote the difficulty level. Additionally, for the WOD-DA, only vehicle and pedestrian classes have labeled annotations for evaluation.

\paragraph{Implementation and training details.}
We trained models using two LiDAR returns and a single frame. The specific implementation and training details for each baseline are provided in the supplementary material. In the DRET, we used atmospheric parameters following \cite{fogsim} and \cite{LISA}. In the SRKD framework, we set the hyper-parameters $\eta_1$, $\eta_2$, and $\eta_3$ (eq.\ref{eq_all}) as 2.0, 0.5, and 2.0.

\subsection{Main Experiment Results}
We conducted three model comparison groups. The models were trained on the WOD-P training set based on DRET augmentation (DRET-Aug) and evaluated on the WOD-P validation set and the rainy data of the WOD-DA.

\paragraph{Performance comparison under rainy weather.}
We first evaluated rainy performance on the WOD-DA. Table \ref{table:waymorain} shows our proposed DRET-Aug and SRKD significantly improve the performance of all 3D detectors in rain. With the SRKD training using DRET-Aug augmentation, Voxel-RCNN, PV-RCNN++, and DSVT achieved improvements of 4.3\%, 3.0\%, and 3.2\% on All(L2-mAP), respectively. Notably, all detectors show substantial gains in the pedestrian class, particularly Voxel-RCNN with a remarkable improvement of 6.5\% on Ped.(L2-mAP). This demonstrates the capability of our framework to improve challenging small object detection under rainy conditions. The results demonstrate the effectiveness and broad applicability of our design.

\paragraph{Performance comparison under sunny weather.}
In contrast to existing methods \cite{stf,snowod} that only focus on performance under adverse weather, it is also essential to maintain performance in sunny weather. Table \ref{table:waymosun} demonstrates that our framework also slightly improves the performance of 3D detectors in sunny conditions. The performance gains mainly come from the enhanced robustness to handle sparse and indistinguishable objects.
%\subsection{Comparison with State-of-the-art}
\paragraph{Rain simulation methods.}
In Table \ref{table:simsota}, we further compared our DRET with the leading rain simulation methods, LISA-Aug and SPRAY-Aug. We observed that plain rain data augmentation alone does not provide significant improvement, as mentioned earlier. Although SPRAY-Aug achieves the best performance on All(L1-mAP), slightly outperforming our DRET-Aug. However, combining all metrics, our method still performs better, especially at the more comprehensive L2 difficulty. Additionally, our DRET-Aug achieves an improvement of 0.5\% and 0.6\% on All(L2-mAP) compared to LISA-Aug and SPRAY-Aug. These results again prove DRET's realism by analyzing the rainy performance gained from simulated data.

\begin{table}[!t]
	\centering
 \resizebox{\columnwidth}{!}{
	\begin{tabular}{c|c|c}
		\hline
		\multirow{2}*{Methods}   &{All(L1)} &{All(L2)}\\             
                                       &mAP/mAPH   &mAP/mAPH
                                       \\
        \hline               
        Baseline
        &\large65.3/62.6  &\large55.8/53.7\\
        + LISA-Aug \cite{LISA}
        &\large65.5/63.2  &\large56.1/54.3\\
        + SPRAY-Aug \cite{spray}
        &\large\textbf{66.0}/63.0  &\large56.0/53.9\\
        
        + DRET-Aug (Ours)
        &\large65.9/\textbf{63.4}	&\large\textbf{56.6}/\textbf{54.7}\\
  \hline
	\end{tabular}
 }
    \caption{Comparison with state-of-the-art rain simulation methods on the WOD-DA.(100\% training data).}
    \label{table:simsota}
    
\end{table}

\paragraph{Robust 3D object detection methods in rain.}
Given the limited prior work on 3D object detection under rainy conditions, we compared our method to the de-noising \cite{cnn_denoise} and  SPG \cite{spg}. As shown in Table \ref{table:spg}, our SRKD has improved by 4.0\% and 2.1\% All(L2-mAP) respectively compared to De-Noising and SPG.

\subsection{Ablation Study}
\iffalse
 To examine the effectiveness of each component in our method, we conducted a series of ablation experiments.
\fi
\begin{table}[!t]
	\centering
 \resizebox{\columnwidth}{!}{
	\begin{tabular}{c|c|c}
		\hline
		\multirow{2}*{Methods}   &{All(L1)} &{All(L2)}\\             
                                       &mAP/mAPH   &mAP/mAPH
                                       \\
        \hline               
        Baseline
        &\large65.3/62.6  &\large55.8/53.7\\
        + De-Noising \cite{cnn_denoise}
        &\large66.9/64.1  &\large56.1/55.4\\
        + SPG \cite{spg}
        &\large67.5/64.7  &\large58.0/55.9\\
        
        + SRKD (Ours)
        &\large\textbf{69.5}/\textbf{66.2}	&\large\textbf{60.1}/\textbf{57.3}\\
  \hline
	\end{tabular}
 }
    \caption{Comparison with state-of-the-art 3D object detection in rain methods on the WOD-DA. (100\% training data).}
    \label{table:spg}
    
\end{table}

\begin{table}[!t]
	\centering
 \resizebox{\columnwidth}{!}{
	\begin{tabular}{c|c|c}
		\hline
		\multirow{2}*{Methods}   &{All(L1)} &{All(L2)}\\             
                                       &mAP/mAPH   &mAP/mAPH
                                       \\
        \hline               
        Voxel-RCNN \cite{voxelrcnn}
        &65.3/62.6  &55.8/53.7\\
        + Our DRET-Aug
        &\textbf{65.9}/\textbf{63.4}	&\textbf{56.6}/\textbf{54.7}\\
        \iffalse
        \rowcolor{LightCyan}
        \textit{Improvement}
        
        &\textit{+0.6}/\textit{+0.8}	&\textit{+0.8}/\textit{+1.0}
        
        \\
        \fi
    \hline
        
        PV-RCNN++ \cite{pvrcnn++}
        &66.2/63.4	&56.8/54.5
        \\

        + Our DRET-Aug
        &\textbf{67.0}/\textbf{64.1}	&\textbf{57.8}/\textbf{55.4}
        \\
        \iffalse
        \rowcolor{LightCyan}
        \textit{Improvement}

        &\textit{+0.8}/\textit{+0.6}	&\textit{+1.0}/\textit{+0.9}	
        \\
        \fi
    \hline
        
        DSVT \cite{dsvt}
        &{69.3}/{66.5}	&{60.4}/{57.9}
        \\
        
        + Our DRET-Aug
         &\textbf{69.9}/\textbf{67.3} &\textbf{61.0}/\textbf{58.7}	\\
        \iffalse
        \rowcolor{LightCyan}
        
        \textit{Improvement}
 
        &\textit{+0.6}/\textit{+0.8}	&\textit{+0.6}/\textit{+0.8}	
        \\	
        \fi	
  \hline
	\end{tabular}
 }
    \caption{The rainy testing results of 3D detectors with our proposed DRET-Aug on the WOD-DA (100\% training data).}	
    \label{table:dret}
\end{table}

\paragraph{Data augmentation based on DRET-Aug.}
To further explore the improvements achieved solely by our DRET-Aug data augmentation, we trained the above baseline 3D detectors with DRET-Aug alone. As shown in Table \ref{table:dret}, DRET-Aug improves Voxel-RCNN, PV-RCNN++ and DSVT by 0.8\%, 1.0\% and 0.6\% on All(L2-mAP), respectively.

\paragraph{Component of SRKD analysis.}

Furthermore, to validate the efficacy of each component in our proposed SRKD, we conducted an ablation study using 20\% training data. The baseline model used is Voxel-RCNN \cite{voxelrcnn} trained with our DRET-Aug. As shown in Table \ref{table:abla}, the inclusion of our proposed components AWID, PRD, and NAPC significantly improves the performance of the Voxel-RCNN baseline. Specifically, AWID, PRD, and NAPC yield gains of 2.0\%, 0.9\%, and 0.7\% on All(L1-mAP), respectively. This highlights the effectiveness of our SRKD components.

\paragraph{Similarity weighting strategies.}

Finally, we tested the different weighting strategies in our AWID module using shape ($\hat{\mathcal{S}}$) and density ($\tilde{\mathcal{S}}$) similarity. As shown in Table \ref{table:abla_Simi}, directly distilling rainy-sunny instance features without any weighting performs poorly. This confirms that it is not reasonable to distill all instances equally.
Moreover, using both $\hat{\mathcal{S}}$ and $\tilde{\mathcal{S}}$ as weighting strategies yields the best performance, improving upon the no weighting strategy by 2.5\% on All (L1-mAP). This validates the importance of our designed adaptive weighting for distilling instance features.

\begin{table}[!t]

    \small
	\centering
	\begin{tabular}{c | c c c  |  c}
		\hline
		\multirow{2}*{Method} &\multirow{2}*{AWID}  &\multirow{2}*{PRD}       &\multirow{2}*{NAPC}   &{All(L1)}  \\
        & & & &mAP/mAPH  \\
		\hline
        \multirow{4}*{Voxel-RCNN} 
         &  &  &  &63.3/60.7       \\
        &\Checkmark &  &  &65.3/62.7 \\
        &\Checkmark &\Checkmark &  &66.2/63.7  \\
        &\Checkmark &\Checkmark &\Checkmark &\textbf{66.9/64.2} \\

		\hline
	\end{tabular}
    \caption{Effects of the different components of SRKD. }

    \label{table:abla}
\end{table}

\begin{table}[!t]

    \small
	\centering
	\begin{tabular}{c |  c c  |  c}
		\hline
		\multirow{2}*{Method}  &\multirow{2}*{$\hat{\mathcal{S}}$}   &\multirow{2}*{$\tilde{\mathcal{S}}$}      &{All(L1)}  \\
        &  &  &mAP/mAPH  \\
		\hline
        \multirow{5}*{Voxel-RCNN + AWID} 
        & &     &62.8/60.2     \\
        &\Checkmark   &     &64.0/61.4 \\
        &   &\Checkmark    &64.8/62.3  \\
        &\Checkmark   &\Checkmark  &\textbf{65.3/62.7} \\
        
		\hline
	\end{tabular}
    \caption{Effects of the different weighting strategy. }

    \label{table:abla_Simi}
\end{table}

\section{Conclusion}
We propose a novel design including DRET and SRKD for 3D detection in sunny and rainy conditions. It uses DRET, a realistic rain simulation, to generate rainy data and mitigate data scarcity. Our SRKD framework then transfers sunny knowledge to rainy detectors via sunny-to-rainy knowledge distillation. Sufficient experimental results of the WOD show that our design can enhance the robustness of 3D detectors in rain. Moreover, our design can even slightly improve the sunny performance of 3D detectors.

\paragraph{Limitations.} Our two-stage DRET requires preprocessing to generate particle sets and cannot be end-to-end. While inference efficiency is unchanged, training time increases unavoidably mainly due to shape similarity calculation.

\section{Acknowledgements} This work was supported in part by the National Natural Science Foundation of China (No.62171393), and the Fundamental Research Funds for the Central Universities (No.20720220064).

\section{SRKD: Supplementary Material}
\subsection{Detailed Visualization of Real Rainy Point Clouds}
\begin{figure}[!h]
\centering
\includegraphics[width=\columnwidth]{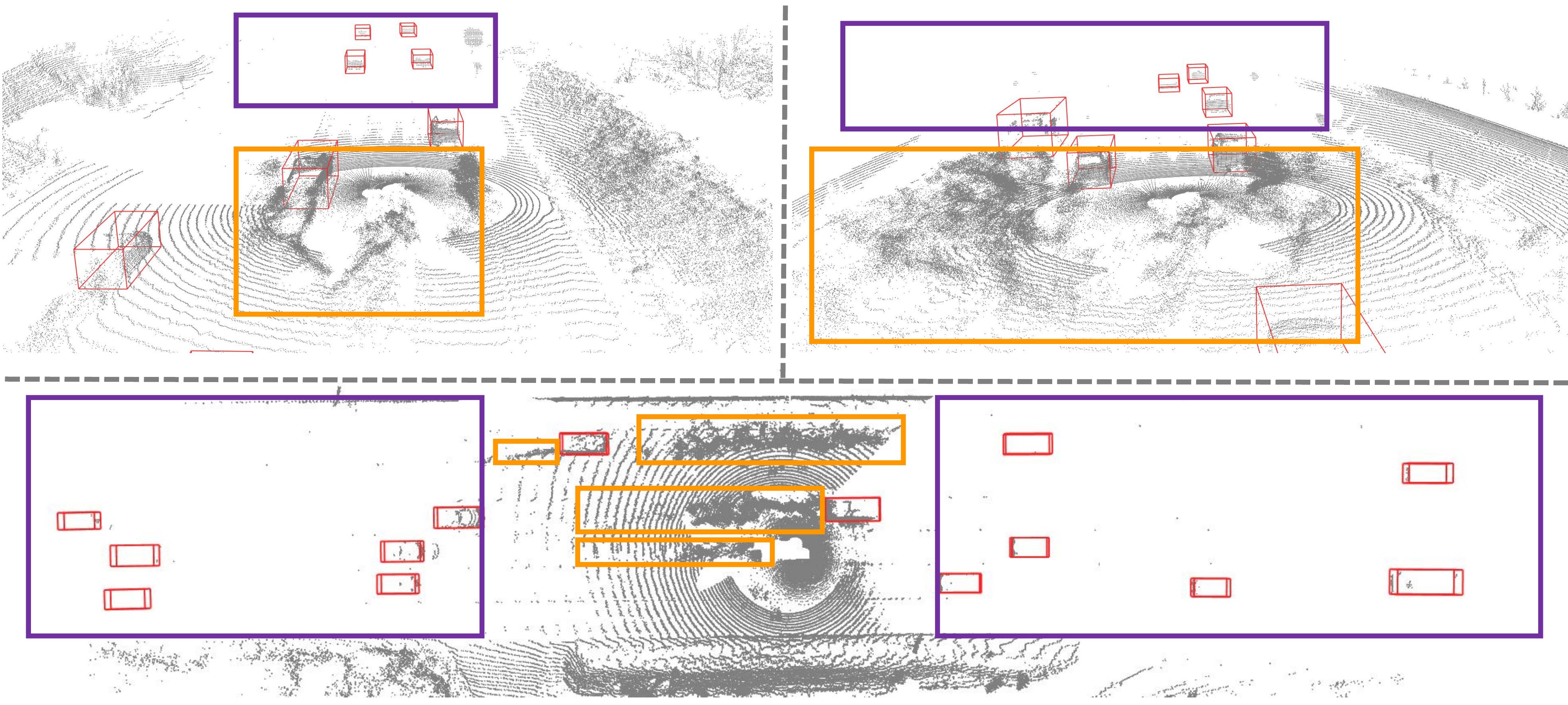} % Reduce the figure size so that it is slightly narrower than the column. Don't use precise values for figure width.This setup will avoid overfull boxes.
\caption{Visualizations of three real rainy day point clouds. Orange is the region where dense rain noise appears, purple is the region where vanishing points appear. }
\label{rain}
\end{figure}

The main paper mentions \emph{dense rain noise} and \emph{missing points} two critical phenomena in rainy point clouds. In order to understand these two phenomena more obviously, we visualize three rainy point clouds in WOD-DA as shown in Fig.~\ref{rain}. Meanwhile, we box the region where the \emph{dense rain noise} occurs in orange and the region where the \emph{missing points} occurs in purple.

\subsection{Detailed Analysis of Performance Improvement}

\begin{figure}[!h]
\centering
\includegraphics[width=\columnwidth]{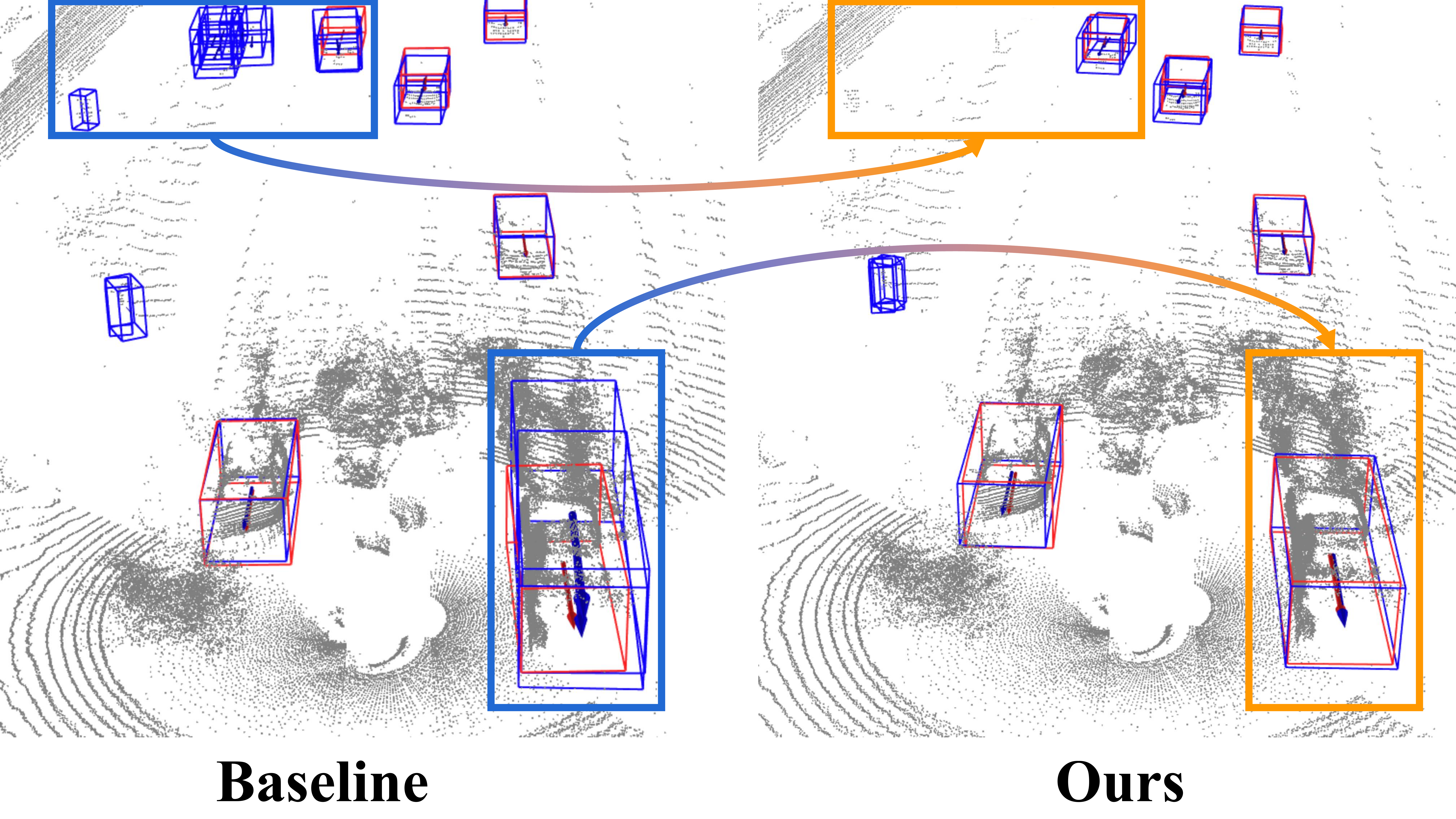} % Reduce the figure size so that it is slightly narrower than the column. Don't use precise values for figure width.This setup will avoid overfull boxes.
\caption{Visualization comparison of detection results in rain. Red boxes indicate groundtruth, and blue boxes indicate test results.}
\label{det}
\end{figure}

As discussed previously in the main paper, rainy conditions significantly decrease the accuracy and recall of 3D detectors. To investigate whether our proposed DRET-Aug and SRKD methods solve this issue, we conducted the same testing as Table 1 in paper on Voxel-RCNN using our proposed method. The results in Table \ref{table:ours pre+recall} show that using the proposed DRET-Aug and SRKD leads to significant improvements across all metrics at various IoU thresholds. Specifically, DRET-Aug and SKRD improve Voxel-RCNN 5.6\% in precision and 2.9\% in recall while the IoU threshold is 0.5 and 0.7, respectively. These experiments demonstrate that our DRET-Aug and SRKD effectively reduce the false positives and negatives that enhance 3D object detection in rain. This result can likewise be further observed in Fig.~\ref{det}, where it can be found that the robustness of the baseline 3D detector in rain is significantly enhanced by adding our method.

\begin{table}[!h]

    \small
	\centering
    \resizebox{\columnwidth}{!}{
	\begin{tabular}{c| c  c   c  | c   c  c }
		\hline
		\multirow{2}*{Method}  &\multicolumn{3}{c|}{Precision}   &\multicolumn{3}{c}{Recall} \\  
		  &0.3 &0.5 &0.7  &0.3 &0.5 &0.7\\
		\hline
		Voxel-RCNN  &33.6	&32.3	&27.3	&83.2	&80.0	&66.7\\
        +Ours    &{38.0} &{36.7} &{30.9} &{85.8} &{82.1} &{69.6}\\
        
        \textit{Improvement}     &\textit{+5.6}	&\textit{+4.3} &\textit{+3.6} &\textit{+2.6} &\textit{+2.1} &\textit{+2.9}\\
		\hline
	\end{tabular}
    }
    \caption{Comparison of vehicle precision and recall under rainy weather. 0.3, 0.5, and 0.7 are the IoU thresholds. }
    \label{table:ours pre+recall}
\end{table}

Table \ref{table:ours pre+recall} in shows the precision and recall improvements by our method, intended to echo our motivation.
Following the reviewer's suggestion, we also conducted more detailed ablation studies to further demonstrate the effectiveness of AWID and NAPC.
The results are shown below (20\% training data, $IoU$ threshold is 0.3):
\begin{table}[!h]

\centering
\resizebox{1\columnwidth}{!}{
                            \begin{tabular}{ccc||ccc}
                \hline
                Model  & FP  & Precision &Model  & FN  & Recall \\ \hline
                Voxel-RCNN  & 70.5k  & 31.9  &Voxel-RCNN &8.1k & 82.0  \\
                + Our NAPC  & 57.8k  & 37.1 &+ Our AWID  &7.2k & 84.6  \\
                \rowcolor{LightCyan}
                \textit{Improvement} & \textit{$\downarrow$18.0\%}  & \textit{$\uparrow$5.2}
                &\textit{Improvement} & \textit{$\downarrow$11.1\%}  & \textit{$\uparrow$2.6}\\
                \hline
                \end{tabular}
                }
\end{table}

\subsection{Implementation and Training Details}
\begin{table}[!ht]
	\centering
 \resizebox{1\columnwidth}{!}{
	\begin{tabular}{c|c |c | c | c | c}
		\hline
		Student    &Teacher &GPUs    &Total Batchsize &Sync\_BN &Epoch \\ 
        \hline               
        Voxel-RCNN
        &Voxel-RCNN	&4×RTX 3090	&4×6=24	&\XSolidBrush	&40 \\
        
    \hline
        
        PV-RCNN++
        &PV-RCNN++	&4×RTX 3090	&4×4=16	&\XSolidBrush	&40	\\
        
    \hline
        
        DSVT
        &DSVT	&6×RTX 3090	&6×3=18	&\Checkmark	&30 \\
    
    \hline
	\end{tabular}
 }
    \caption{Implementation and training details of three 3D detectors with our DRET-Aug and SRKD.}	
    
    \label{table:detail}
\end{table}
Table \ref{table:detail} is a detailed supplement to the main paper experimental implementation and training details. The code implementation is based on the open-source code of the corresponding 3D detector in the OpenPCDet project. 
Additionally, the rest of the training details are shown in Table~\ref{table:detail}.

\subsection{Computational and Memory Cost}

\noindent 
We tested the single-stage DSVT and the two-stage PV-RCNN++. Below, we present the computational and memory costs incurred during the \textbf{training phase}. 
%We selected the {single stage} DSVT and the {two stage} PV-RCNN++ for testing. The computational and memory costs during the \textbf{training phase} are shown below.

\begin{table}[!h]

\centering
\resizebox{\columnwidth}{!}{
    \begin{tabular}{c|c|c} 
    
    \hline  
    Model & Training time & Memory\\
    \hline
    \hline
    DSVT & 49h &21G\\
    + Our SRKD & 61h & 23G\\
    
    \rowcolor{LightCyan}
    Cost of SRKD in DSVT & $\uparrow$24\% &  $\uparrow$10\% \\
    \hline 
    PV-RCNN++ & 54h &21G\\
    + Our SRKD& 83h & 24G\\
    
    \rowcolor{LightCyan}
    Cost of SRKD in PV-RCNN++ & $\uparrow$54\% &  $\uparrow$14\% \\
    \hline
    \end{tabular} 
}
\end{table}
\noindent 
The results indicate our SRKD method does incur additional computational and memory costs, but the extra costs are not substantial because the teacher model does not require gradient computation and memory optimization. 
Furthermore, during the \textbf{inference phase}, as we only use the student model and do not calculate the loss, there are no additional computational and memory costs. 
%The results show that our proposed SRKD does introduce additional computational and memory costs. However, since the teacher model does not require gradient computation and memory optimization techniques, \textbf{the extra cost is not substantial}. Furthermore, during \textbf{inference phase}, as we only use the student model and do not need to calculate the loss, {there is no extra computational and memory cost}.

% Generated by IEEEtran.bst, version: 1.14 (2015/08/26)


\begin{thebibliography}{29}
\providecommand{\natexlab}[1]{#1}

\bibitem[{Bijelic et~al.(2020)Bijelic, Gruber, Mannan, Kraus, Ritter,
  Dietmayer, and Heide}]{stf}
Bijelic, M.; Gruber, T.; Mannan, F.; Kraus, F.; Ritter, W.; Dietmayer, K.; and
  Heide, F. 2020.
\newblock Seeing {{Through Fog Without Seeing Fog}}: {{Deep Multimodal Sensor
  Fusion}} in {{Unseen Adverse Weather}}.
\newblock In \emph{{{CVPR}}}.

\bibitem[{Borgefors(1986)}]{charmfer}
Borgefors, G. 1986.
\newblock Distance transformations in digital images.
\newblock \emph{Computer vision, graphics, and image processing}, 34(3):
  344--371.

\bibitem[{Bridson, Houriham, and Nordenstam(2007)}]{perlin}
Bridson, R.; Houriham, J.; and Nordenstam, M. 2007.
\newblock Curl-Noise for Procedural Fluid Flow.
\newblock \emph{ACM Trans. Graph.}, 26(3): 46–es.

\bibitem[{Charron, Phillips, and Waslander(2018)}]{denoise_snow}
Charron, N.; Phillips, S.; and Waslander, S.~L. 2018.
\newblock De-Noising of {{Lidar Point Clouds Corrupted}} by {{Snowfall}}.
\newblock In \emph{{{CRV}}}.

\bibitem[{Deng et~al.(2021)Deng, Shi, Li, Zhou, Zhang, and Li}]{voxelrcnn}
Deng, J.; Shi, S.; Li, P.; Zhou, W.; Zhang, Y.; and Li, H. 2021.
\newblock Voxel {{R-CNN}}: {{Towards High Performance Voxel-based 3D Object
  Detection}}.
\newblock \emph{AAAI}, 35.

\bibitem[{Do and Yoo(2022)}]{sr}
Do, A.~T.; and Yoo, M. 2022.
\newblock LossDistillNet: 3D Object Detection in Point Cloud Under Harsh
  Weather Conditions.
\newblock \emph{IEEE Access}, 10: 84882--84893.

\bibitem[{Hahner et~al.(2022)Hahner, Sakaridis, Bijelic, Heide, Yu, Dai, and
  Van~Gool}]{snowsim}
Hahner, M.; Sakaridis, C.; Bijelic, M.; Heide, F.; Yu, F.; Dai, D.; and
  Van~Gool, L. 2022.
\newblock {{LiDAR Snowfall Simulation}} for {{Robust 3D Object Detection}}.
\newblock In \emph{{{CVPR}}}.

\bibitem[{Hahner et~al.(2021)Hahner, Sakaridis, Dai, and Van~Gool}]{fogsim}
Hahner, M.; Sakaridis, C.; Dai, D.; and Van~Gool, L. 2021.
\newblock Fog {{Simulation}} on {{Real LiDAR Point Clouds}} for {{3D Object
  Detection}} in {{Adverse Weather}}.
\newblock In \emph{{{ICCV}}}.

\bibitem[{He et~al.(2020)He, Zeng, Huang, Hua, and Zhang}]{sassd}
He, C.; Zeng, H.; Huang, J.; Hua, X.-S.; and Zhang, L. 2020.
\newblock Structure {{Aware Single-Stage 3D Object Detection From Point
  Cloud}}.
\newblock In \emph{{{CVPR}}}.

\bibitem[{Heinzler et~al.(2020)Heinzler, Piewak, Schindler, and
  Stork}]{cnn_denoise}
Heinzler, R.; Piewak, F.; Schindler, P.; and Stork, W. 2020.
\newblock {{CNN-Based Lidar Point Cloud De-Noising}} in {{Adverse Weather}}.
\newblock \emph{IEEE Robotics and Automation Letters}, 5.

\bibitem[{Kilic et~al.(2021)Kilic, Hegde, Sindagi, Cooper, Foster, and
  Patel}]{LISA}
Kilic, V.; Hegde, D.; Sindagi, V.~A.; Cooper, A.; Foster, M.; and Patel, V.~M.
  2021.
\newblock Lidar {{Light Scattering Augmentation}} ({{LISA}}): {{Physics-based
  Simulation}} of {{Adverse Weather Conditions}} for {{3D Object Detection}}.
\newblock \emph{ArXiv}.

\bibitem[{Lin et~al.(2022)Lin, Yin, Yan, Ge, Zhang, and Rigoll}]{snowod}
Lin, J.; Yin, H.; Yan, J.; Ge, W.; Zhang, H.; and Rigoll, G. 2022.
\newblock Improved {{3D Object Detector Under Snowfall Weather Condition
  Based}} on {{LiDAR Point Cloud}}.
\newblock \emph{IEEE Sensors Journal}, 22.

\bibitem[{Sheng et~al.(2021)Sheng, Cai, Liu, Deng, Huang, Hua, and Zhao}]{ct3d}
Sheng, H.; Cai, S.; Liu, Y.; Deng, B.; Huang, J.; Hua, X.-S.; and Zhao, M.-J.
  2021.
\newblock Improving {{3D Object Detection With Channel-Wise Transformer}}.
\newblock In \emph{{{ICCV}}}.

\bibitem[{Shi et~al.(2020)Shi, Guo, Jiang, Wang, Shi, Wang, and Li}]{pvrcnn}
Shi, S.; Guo, C.; Jiang, L.; Wang, Z.; Shi, J.; Wang, X.; and Li, H. 2020.
\newblock {{PV-RCNN}}: {{Point-Voxel Feature Set Abstraction}} for {{3D Object
  Detection}}.
\newblock In \emph{{{CVPR}}}.

\bibitem[{Shi et~al.(2022)Shi, Jiang, Deng, Wang, Guo, Shi, Wang, and
  Li}]{pvrcnn++}
Shi, S.; Jiang, L.; Deng, J.; Wang, Z.; Guo, C.; Shi, J.; Wang, X.; and Li, H.
  2022.
\newblock {{PV-RCNN}}++: {{Point-Voxel Feature Set Abstraction With Local
  Vector Representation}} for {{3D Object Detection}}.
\newblock \emph{Int. J. Comput. Vision}, 131.

\bibitem[{Shi, Wang, and Li(2019)}]{pointrcnn}
Shi, S.; Wang, X.; and Li, H. 2019.
\newblock {{PointRCNN}}: {{3D Object Proposal Generation}} and {{Detection From
  Point Cloud}}.
\newblock In \emph{{{CVPR}}}.

\bibitem[{Shih et~al.(2022)Shih, Liao, Lin, Wong, and Wang}]{spray}
Shih, Y.-C.; Liao, W.-H.; Lin, W.-C.; Wong, S.-K.; and Wang, C.-C. 2022.
\newblock Reconstruction and {{Synthesis}} of {{Lidar Point Clouds}} of
  {{Spray}}.
\newblock \emph{IEEE Robotics and Automation Letters}, 7.

\bibitem[{Sun et~al.(2020)Sun, Kretzschmar, Dotiwalla, Chouard, Patnaik, Tsui,
  Guo, Zhou, Chai, Caine, Vasudevan, Han, Ngiam, Zhao, Timofeev, Ettinger,
  Krivokon, Gao, Joshi, Zhang, Shlens, Chen, and Anguelov}]{waymo}
Sun, P.; Kretzschmar, H.; Dotiwalla, X.; Chouard, A.; Patnaik, V.; Tsui, P.;
  Guo, J.; Zhou, Y.; Chai, Y.; Caine, B.; Vasudevan, V.; Han, W.; Ngiam, J.;
  Zhao, H.; Timofeev, A.; Ettinger, S.; Krivokon, M.; Gao, A.; Joshi, A.;
  Zhang, Y.; Shlens, J.; Chen, Z.; and Anguelov, D. 2020.
\newblock Scalability in {{Perception}} for {{Autonomous Driving}}: {{Waymo
  Open Dataset}}.
\newblock In \emph{{{CVPR}}}.

\bibitem[{Teufel et~al.(2022)Teufel, Volk, Von~Bernuth, and Bringmann}]{sim1}
Teufel, S.; Volk, G.; Von~Bernuth, A.; and Bringmann, O. 2022.
\newblock Simulating {{Realistic Rain}}, {{Snow}}, and {{Fog Variations For
  Comprehensive Performance Characterization}} of {{LiDAR Perception}}.
\newblock In \emph{2022 {{IEEE}} 95th {{Vehicular Technology Conference}}:
  ({{VTC2022-Spring}})}.

\bibitem[{Wang et~al.(2023)Wang, Shi, Shi, Lei, Wang, He, Schiele, and
  Wang}]{dsvt}
Wang, H.; Shi, C.; Shi, S.; Lei, M.; Wang, S.; He, D.; Schiele, B.; and Wang,
  L. 2023.
\newblock {{DSVT}}: {{Dynamic Sparse Voxel Transformer With Rotated Sets}}.
\newblock In \emph{{{CVPR}}}.

\bibitem[{Wu et~al.(2023)Wu, Wen, Shi, Li, and Wang}]{virconv}
Wu, H.; Wen, C.; Shi, S.; Li, X.; and Wang, C. 2023.
\newblock Virtual {{Sparse Convolution}} for {{Multimodal 3D Object
  Detection}}.
\newblock In \emph{{{CVPR}}}.

\bibitem[{Wu et~al.(2022)Wu, Peng, Yang, Xie, Huang, Deng, Liu, and Cai}]{sfd}
Wu, X.; Peng, L.; Yang, H.; Xie, L.; Huang, C.; Deng, C.; Liu, H.; and Cai, D.
  2022.
\newblock Sparse {{Fuse Dense}}: {{Towards High Quality 3D Detection}} with
  {{Depth Completion}}.
\newblock In \emph{{{CVPR}}}.

\bibitem[{Xu et~al.(2021)Xu, Zhou, Wang, Qi, and Anguelov}]{spg}
Xu, Q.; Zhou, Y.; Wang, W.; Qi, C.~R.; and Anguelov, D. 2021.
\newblock {{SPG}}: {{Unsupervised Domain Adaptation}} for {{3D Object
  Detection}} via {{Semantic Point Generation}}.
\newblock In \emph{{{ICCV}}}.

\bibitem[{Yan, Mao, and Li(2018)}]{second}
Yan, Y.; Mao, Y.; and Li, B. 2018.
\newblock {{SECOND}}: {{Sparsely Embedded Convolutional Detection}}.
\newblock \emph{Sensors}, 18.

\bibitem[{Yang et~al.(2022)Yang, Shi, Ding, Wang, and Qi}]{spareKD}
Yang, J.; Shi, S.; Ding, R.; Wang, Z.; and Qi, X. 2022.
\newblock Towards {{Efficient 3D Object Detection}} with {{Knowledge
  Distillation}}.
\newblock \emph{NIPS}, 35.

\bibitem[{Yang et~al.(2019)Yang, Sun, Liu, Shen, and Jia}]{std3d}
Yang, Z.; Sun, Y.; Liu, S.; Shen, X.; and Jia, J. 2019.
\newblock {{STD}}: {{Sparse-to-Dense 3D Object Detector}} for {{Point Cloud}}.
\newblock In \emph{{{ICCV}}}.

\bibitem[{Zheng et~al.(2022{\natexlab{a}})Zheng, Hong, Jiang, and Fu}]{boost1}
Zheng, W.; Hong, M.; Jiang, L.; and Fu, C.-W. 2022{\natexlab{a}}.
\newblock Boosting {{3D Object Detection}} by {{Simulating Multimodality}} on
  {{Point Clouds}}.
\newblock In \emph{{{CVPR}}}.

\bibitem[{Zheng et~al.(2022{\natexlab{b}})Zheng, Jiang, Lu, Ye, and
  Fu}]{boost2}
Zheng, W.; Jiang, L.; Lu, F.; Ye, Y.; and Fu, C.-W. 2022{\natexlab{b}}.
\newblock Boosting {{Single-Frame 3D Object Detection}} by {{Simulating
  Multi-Frame Point Clouds}}.
\newblock In \emph{ACMMM}.

\bibitem[{Zheng et~al.(2021)Zheng, Tang, Jiang, and Fu}]{sessd}
Zheng, W.; Tang, W.; Jiang, L.; and Fu, C.-W. 2021.
\newblock {{SE-SSD}}: {{Self-Ensembling Single-Stage Object Detector From Point
  Cloud}}.
\newblock In \emph{{{CVPR}}}.

\end{thebibliography}


% Generated by IEEEtran.bst, version: 1.14 (2015/08/26)
\begin{thebibliography}{10}
\providecommand{\url}[1]{#1}
\csname url@samestyle\endcsname
\providecommand{\newblock}{\relax}
\providecommand{\bibinfo}[2]{#2}
\providecommand{\BIBentrySTDinterwordspacing}{\spaceskip=0pt\relax}
\providecommand{\BIBentryALTinterwordstretchfactor}{4}
\providecommand{\BIBentryALTinterwordspacing}{\spaceskip=\fontdimen2\font plus
\BIBentryALTinterwordstretchfactor\fontdimen3\font minus \fontdimen4\font\relax}
\providecommand{\BIBforeignlanguage}[2]{{%
\expandafter\ifx\csname l@#1\endcsname\relax
\typeout{** WARNING: IEEEtran.bst: No hyphenation pattern has been}%
\typeout{** loaded for the language `#1'. Using the pattern for}%
\typeout{** the default language instead.}%
\else
\language=\csname l@#1\endcsname
\fi
#2}}
\providecommand{\BIBdecl}{\relax}
\BIBdecl

\bibitem{3dssd}
Z.~Yang, Y.~Sun, S.~Liu, and J.~Jia, ``{{3DSSD}}: {{Point-Based 3D Single Stage Object Detector}},'' in \emph{{{CVPR}}}, 2020.

\bibitem{addataset1}
X.~Huang, P.~Wang, X.~Cheng, D.~Zhou, Q.~Geng, and R.~Yang, ``The {{ApolloScape Open Dataset}} for {{Autonomous Driving}} and its {{Application}},'' \emph{IEEE Trans. Pattern Anal. Mach. Intell.}, vol.~42, 2020.

\bibitem{addataset3}
C.~A. {Diaz-Ruiz}, Y.~Xia, Y.~You, J.~Nino, J.~Chen, J.~Monica, X.~Chen, K.~Luo, Y.~Wang, M.~Emond, W.-L. Chao, B.~Hariharan, K.~Q. Weinberger, and M.~Campbell, ``Ithaca365: {{Dataset}} and {{Driving Perception Under Repeated}} and {{Challenging Weather Conditions}},'' in \emph{{{CVPR}}}, 2022.

\bibitem{adddataset2}
M.~Pitropov, D.~Garcia, J.~Rebello, M.~Smart, C.~Wang, K.~Czarnecki, and S.~Waslander, ``Canadian {{Adverse Driving Conditions Dataset}},'' \emph{The International Journal of Robotics Research}, vol.~40, 2021.

\bibitem{boost1}
W.~Zheng, M.~Hong, L.~Jiang, and C.-W. Fu, ``Boosting {{3D Object Detection}} by {{Simulating Multimodality}} on {{Point Clouds}},'' in \emph{{{CVPR}}}, 2022.

\bibitem{boost2}
W.~Zheng, L.~Jiang, F.~Lu, Y.~Ye, and C.-W. Fu, ``Boosting {{Single-Frame 3D Object Detection}} by {{Simulating Multi-Frame Point Clouds}},'' in \emph{ACMMM}, 2022.

\bibitem{cnn_denoise}
R.~Heinzler, F.~Piewak, P.~Schindler, and W.~Stork, ``{{CNN-Based Lidar Point Cloud De-Noising}} in {{Adverse Weather}},'' \emph{IEEE Robotics and Automation Letters}, vol.~5, 2020.

\bibitem{ct3d}
H.~Sheng, S.~Cai, Y.~Liu, B.~Deng, J.~Huang, X.-S. Hua, and M.-J. Zhao, ``Improving {{3D Object Detection With Channel-Wise Transformer}},'' in \emph{{{ICCV}}}, 2021.

\bibitem{denoise_snow}
N.~Charron, S.~Phillips, and S.~L. Waslander, ``De-noising of {{Lidar Point Clouds Corrupted}} by {{Snowfall}},'' in \emph{{{CRV}}}, 2018.

\bibitem{dsvt}
H.~Wang, C.~Shi, S.~Shi, M.~Lei, S.~Wang, D.~He, B.~Schiele, and L.~Wang, ``{{DSVT}}: {{Dynamic Sparse Voxel Transformer With Rotated Sets}},'' in \emph{{{CVPR}}}, 2023.

\bibitem{fogsim}
M.~Hahner, C.~Sakaridis, D.~Dai, and L.~Van~Gool, ``Fog {{Simulation}} on {{Real LiDAR Point Clouds}} for {{3D Object Detection}} in {{Adverse Weather}},'' in \emph{{{ICCV}}}, 2021.

\bibitem{iassd}
Y.~Zhang, Q.~Hu, G.~Xu, Y.~Ma, J.~Wan, and Y.~Guo, ``Not {{All Points Are Equal}}: {{Learning Highly Efficient Point-Based Detectors}} for {{3D LiDAR Point Clouds}},'' in \emph{{{CVPR}}}, 2022.

\bibitem{LISA}
V.~Kilic, D.~Hegde, V.~A. Sindagi, A.~Cooper, M.~Foster, and V.~M. Patel, ``Lidar {{Light Scattering Augmentation}} ({{LISA}}): {{Physics-based Simulation}} of {{Adverse Weather Conditions}} for {{3D Object Detection}},'' \emph{ArXiv}, 2021.

\bibitem{odcold}
A.~Piroli, V.~Dallabetta, M.~Walessa, D.~Meissner, J.~Kopp, and K.~Dietmayer, ``Robust {{3D Object Detection}} in {{Cold Weather Conditions}},'' in \emph{2022 {{IEEE Intelligent Vehicles Symposium}} ({{IV}})}, 2022.

\bibitem{pointpillars}
A.~H. Lang, S.~Vora, H.~Caesar, L.~Zhou, J.~Yang, and O.~Beijbom, ``{{PointPillars}}: {{Fast Encoders}} for {{Object Detection From Point Clouds}},'' in \emph{{{CVPR}}}, 2019.

\bibitem{pointrcnn}
S.~Shi, X.~Wang, and H.~Li, ``{{PointRCNN}}: {{3D Object Proposal Generation}} and {{Detection From Point Cloud}},'' in \emph{{{CVPR}}}, 2019.

\bibitem{pvrcnn}
S.~Shi, C.~Guo, L.~Jiang, Z.~Wang, J.~Shi, X.~Wang, and H.~Li, ``{{PV-RCNN}}: {{Point-Voxel Feature Set Abstraction}} for {{3D Object Detection}},'' in \emph{{{CVPR}}}, 2020.

\bibitem{pvrcnn++}
S.~Shi, L.~Jiang, J.~Deng, Z.~Wang, C.~Guo, J.~Shi, X.~Wang, and H.~Li, ``{{PV-RCNN}}++: {{Point-Voxel Feature Set Abstraction With Local Vector Representation}} for {{3D Object Detection}},'' \emph{Int. J. Comput. Vision}, vol. 131, 2022.

\bibitem{sassd}
C.~He, H.~Zeng, J.~Huang, X.-S. Hua, and L.~Zhang, ``Structure {{Aware Single-Stage 3D Object Detection From Point Cloud}},'' in \emph{{{CVPR}}}, 2020.

\bibitem{second}
Y.~Yan, Y.~Mao, and B.~Li, ``{{SECOND}}: {{Sparsely Embedded Convolutional Detection}},'' \emph{Sensors}, vol.~18, 2018.

\bibitem{sessd}
W.~Zheng, W.~Tang, L.~Jiang, and C.-W. Fu, ``{{SE-SSD}}: {{Self-Ensembling Single-Stage Object Detector From Point Cloud}},'' in \emph{{{CVPR}}}, 2021.

\bibitem{sfd}
X.~Wu, L.~Peng, H.~Yang, L.~Xie, C.~Huang, C.~Deng, H.~Liu, and D.~Cai, ``Sparse {{Fuse Dense}}: {{Towards High Quality 3D Detection}} with {{Depth Completion}},'' in \emph{{{CVPR}}}, 2022.

\bibitem{sim1}
S.~Teufel, G.~Volk, A.~Von~Bernuth, and O.~Bringmann, ``Simulating {{Realistic Rain}}, {{Snow}}, and {{Fog Variations For Comprehensive Performance Characterization}} of {{LiDAR Perception}},'' in \emph{2022 {{IEEE}} 95th {{Vehicular Technology Conference}}: ({{VTC2022-Spring}})}, 2022.

\bibitem{slsfusion}
N.~A.~M. Mai, P.~Duthon, L.~Khoudour, A.~Crouzil, and S.~A. Velastin, ``{{3D Object Detection}} with {{SLS-Fusion Network}} in {{Foggy Weather Conditions}},'' \emph{Sensors}, vol.~21, 2021.

\bibitem{snowod}
J.~Lin, H.~Yin, J.~Yan, W.~Ge, H.~Zhang, and G.~Rigoll, ``Improved {{3D Object Detector Under Snowfall Weather Condition Based}} on {{LiDAR Point Cloud}},'' \emph{IEEE Sensors Journal}, vol.~22, 2022.

\bibitem{snowsim}
M.~Hahner, C.~Sakaridis, M.~Bijelic, F.~Heide, F.~Yu, D.~Dai, and L.~Van~Gool, ``{{LiDAR Snowfall Simulation}} for {{Robust 3D Object Detection}},'' in \emph{{{CVPR}}}, 2022.

\bibitem{spareKD}
J.~Yang, S.~Shi, R.~Ding, Z.~Wang, and X.~Qi, ``Towards {{Efficient 3D Object Detection}} with {{Knowledge Distillation}},'' \emph{NIPS}, vol.~35, 2022.

\bibitem{spg}
Q.~Xu, Y.~Zhou, W.~Wang, C.~R. Qi, and D.~Anguelov, ``{{SPG}}: {{Unsupervised Domain Adaptation}} for {{3D Object Detection}} via {{Semantic Point Generation}},'' in \emph{{{ICCV}}}, 2021.

\bibitem{spray}
Y.-C. Shih, W.-H. Liao, W.-C. Lin, S.-K. Wong, and C.-C. Wang, ``Reconstruction and {{Synthesis}} of {{Lidar Point Clouds}} of {{Spray}},'' \emph{IEEE Robotics and Automation Letters}, vol.~7, 2022.

\bibitem{std3d}
Z.~Yang, Y.~Sun, S.~Liu, X.~Shen, and J.~Jia, ``{{STD}}: {{Sparse-to-Dense 3D Object Detector}} for {{Point Cloud}},'' in \emph{{{ICCV}}}, 2019.

\bibitem{stf}
M.~Bijelic, T.~Gruber, F.~Mannan, F.~Kraus, W.~Ritter, K.~Dietmayer, and F.~Heide, ``Seeing {{Through Fog Without Seeing Fog}}: {{Deep Multimodal Sensor Fusion}} in {{Unseen Adverse Weather}},'' in \emph{{{CVPR}}}, 2020.

\bibitem{virconv}
H.~Wu, C.~Wen, S.~Shi, X.~Li, and C.~Wang, ``Virtual {{Sparse Convolution}} for {{Multimodal 3D Object Detection}},'' in \emph{{{CVPR}}}, 2023.

\bibitem{voxelrcnn}
J.~Deng, S.~Shi, P.~Li, W.~Zhou, Y.~Zhang, and H.~Li, ``Voxel {{R-CNN}}: {{Towards High Performance Voxel-based 3D Object Detection}},'' \emph{AAAI}, vol.~35, 2021.

\bibitem{waymo}
P.~Sun, H.~Kretzschmar, X.~Dotiwalla, A.~Chouard, V.~Patnaik, P.~Tsui, J.~Guo, Y.~Zhou, Y.~Chai, B.~Caine, V.~Vasudevan, W.~Han, J.~Ngiam, H.~Zhao, A.~Timofeev, S.~Ettinger, M.~Krivokon, A.~Gao, A.~Joshi, Y.~Zhang, J.~Shlens, Z.~Chen, and D.~Anguelov, ``Scalability in {{Perception}} for {{Autonomous Driving}}: {{Waymo Open Dataset}},'' in \emph{{{CVPR}}}, 2020.

\bibitem{perlin}
\BIBentryALTinterwordspacing
R.~Bridson, J.~Houriham, and M.~Nordenstam, ``Curl-noise for procedural fluid flow,'' \emph{ACM Trans. Graph.}, vol.~26, no.~3, p. 46–es, jul 2007. [Online]. Available: \url{https://doi.org/10.1145/1276377.1276435}
\BIBentrySTDinterwordspacing

\bibitem{charmfer}
G.~Borgefors, ``Distance transformations in digital images,'' \emph{Computer vision, graphics, and image processing}, vol.~34, no.~3, pp. 344--371, 1986.

\bibitem{sr}
A.~T. Do and M.~Yoo, ``Lossdistillnet: 3d object detection in point cloud under harsh weather conditions,'' \emph{IEEE Access}, vol.~10, pp. 84\,882--84\,893, 2022.

\bibitem{randlanet}
Q.~Hu, B.~Yang, L.~Xie, S.~Rosa, Y.~Guo, Z.~Wang, N.~Trigoni, and A.~Markham, ``Randla-net: Efficient semantic segmentation of large-scale point clouds,'' in \emph{Proceedings of the IEEE/CVF conference on computer vision and pattern recognition}, 2020, pp. 11\,108--11\,117.

\end{thebibliography}
\end{document}